\documentclass[journal,twocolumn,singlespace]{IEEEtran}
\usepackage{amsbsy,amsmath,amssymb,epsfig,bbm,mathrsfs,multirow,amsthm,subcaption,caption,graphicx,epstopdf,multicol,lipsum,float}
\usepackage{algorithm2e}
\usepackage{adjustbox}
\usepackage{multicol}
\usepackage{enumerate}
\usepackage{tikz}
\usepackage[hidelinks]{hyperref}

\newenvironment{myalign}{\par\nobreak\noindent\align}{\endalign}

\begin{document}

\title{Joint Ground and Aerial Package Delivery Services: A Stochastic Optimization Approach}

\author{Suttinee~Sawadsitang, Dusit~Niyato,~\IEEEmembership{Fellow,~IEEE,} Puay-Siew~Tan, and~Ping~Wang,~\IEEEmembership{Senior~Member,~IEEE}
\thanks{S. Sawadsitang and D. Niyato are with the School of Computer Science and Engineering, Nanyang Technological University, Singapore, 639798, e-mail: (e-mail: suttinee002@e.ntu.edu.sg; dniyato@ntu.edu.sg).}
\thanks{P.-S. Tan is with the Singapore Institute of Manufacturing Technology, Singapore 638075 (e-mail: pstan@simtech.a-star.edu.sg).}
\thanks{P. Wang is with the Department of Electrical Engineering and Computer Science, York University, Canada (e-mail: pingw@yorku.ca).}
\thanks{Color versions of one or more of the figures in this paper are available online
at http://ieeexplore.ieee.org.}}

\maketitle

\begin{abstract}
Unmanned aerial vehicles (UAVs), also known as drones, have emerged as a promising mode of fast, energy-efficient, and cost-effective package delivery. A considerable number of works have studied different aspects of drone package delivery service by a supplier, one of which is delivery planning. However, existing works addressing the planning issues consider a simple case of perfect delivery without service interruption, e.g., due to accident which is common and realistic. Therefore, this paper introduces the joint ground and aerial delivery service optimization and planning (GADOP) framework. The framework explicitly incorporates uncertainty of drone package delivery, i.e., takeoff and breakdown conditions. The GADOP framework aims to minimize the total delivery cost given practical constraints, e.g., traveling distance limit. Specifically, we formulate the GADOP framework as a three-stage stochastic integer programming model. To deal with the high complexity issue of the problem, a decomposition method is adopted. Then, the performance of the GADOP framework is evaluated by using two data sets including Solomon benchmark suite and the real data from one of the Singapore logistics companies. The performance evaluation clearly shows that the GADOP framework can achieve significantly lower total payment than that of the baseline methods which do not take uncertainty into account. 
\end{abstract}


\section{Introduction}

Package delivery becomes an instrumental function of logistics businesses due to the popularity of e-commerce such as online shopping. Traditionally, package delivery has been done through land transportation such as using trucks, cars, motorcycles, and bicycles, which is reasonably affordable, reliable, and accessible. However, due to rising labor costs, an alternative delivery mode by using unmanned aerial vehicles (UAVs) such as drones becomes increasingly attractive. Thanks to the advancement of drone technologies, the cost and reliability of using drones in package delivery services have been improved significantly. Businesses start adopting the drone delivery such as Amazon Prime Air, DHL Parcelcopter, and Google Project Wing, showing the technological and economic feasibility of the services. The benefits of drone package delivery services are palpable as follows. Firstly, drone delivery is faster than land vehicles as drones are not subjected to road traffic jam. Also, it has much less service time as the drones can drop a package directly to the customers with minimal human involved. Secondly, drone delivery reduces resource usage in terms of manpower and energy. Thirdly, drone delivery is applicable to the areas which are difficult to access by land transportation.

Although drone delivery has many advantages over traditional ground-based delivery, using drones still faces many challenges.\footnote{In this paper, ground-based delivery refers to the delivery by trucks.} For example, the reliability of drone delivery is lower than that of the ground-based delivery. Furthermore, drones have significantly less service coverage and capacity. As a result, using the drones for package delivery is limited to a certain area near the depot. Nevertheless, recently, an integration of ground-based delivery and drone delivery has been introduced~\cite{sidekick}. It has been reported that these two delivery modes can complement each other because of their unique features and advantages. For example, packages can be optimally assigned to be delivered by trucks or drones to maximize service quality and to minimize cost. In this regard, a joint ground and aerial delivery service optimization and planning emerge as an important issue. 

In this paper, we introduce the joint ground and aerial delivery service optimization and planning (GADOP) framework for a supplier. In particular, customers' packages can be delivered by trucks or drones. We consider the case that the supplier needs to make reservations for both trucks and drones in advance. The supplier, after deciding packages to be delivered by trucks, determines route for each truck. Similarly, the supplier determines package serving order of the drones. We model the uncertainty of the drone delivery in terms of takeoff and breakdown conditions. Specifically, the drone may not be able to take off, e.g., due to bad weather, and if it can take off, the drone may experience accident, e.g., due to technical issues. Therefore, we adopt the stochastic programming technique to formulate the joint ground and aerial delivery service optimization and planning in which the decisions are made in different stages. The contributions of this paper are summarized as follows.
\begin{itemize}
	\item The proposed joint optimization is able to utilize and trade off benefits and limitations of truck and drone delivery. In particular, the proposed optimization can yield the minimum total payment\footnote{In this paper, total cost and total payment refer to the same.} for the supplier by considering the routing path of trucks and the serving order of drones. 
	\item The formulated stochastic programming model can find and achieve an optimal solution taking the uncertainty of the drone delivery into account. Consequently, the supplier can obtain the better planning for the truck and drone usage given certain constraints such as capacity limit and traveling distance limit. 
	\item To reduce the computational time of obtaining the solution of the stochastic programming model, we introduce the decomposed GADOP by adopting the L-shape method. The decomposition allows the model to be split into multiple smaller sub-problems, which are more computationally tractable.
\end{itemize}
This paper is structured as follows. Section~\ref{sec_related} presents the literature review of vehicle routing problem, ground and aerial delivery. The system model and the formulations of the proposed joint optimization are presented in Section~\ref{sec_system} and Section~\ref{sec_formulation}, respectively. Section~\ref{sec_decom} presents the decomposition of the proposed joint optimization. Section~\ref{sec_eva} presents the experiment setting and the numerical results of the proposed joint optimization. At the end, the conclusion and future works are discussed in Section~\ref{sec_con}. 

\section{Related Work}
\label{sec_related}

Vehicle Routing Problem (VRP) has played a significant role in goods distribution for half century. Hundreds of different models and solution algorithms have been proposed since then. Many surveys on a variety of issues in VRP exist in the literature, e.g.,~\cite{ref_survey_2007-general}, \cite{ref_greensurvey}. Some of them consider specific aspects of the problem, e.g., VRP with time windows~\cite{ref_survey_1988}, pick-up and delivery VRP~\cite{ref_survey_2007-pickup}, dynamic VRP~\cite{ref_survey_2013-dynamic} and stochastic VRP~\cite{ref_survey_1996}, and solution algorithms~\cite{ref_survey_1992}, \cite{ref_survey_2000}. However, almost all of the existing VRP works consider only cars or trucks because they are the most popular mode of transportation. Recently, drones has been introduced which has a great potential to be used in businesses. A number of researchers and engineers studied and improved the technical aspects of drones, i.e., enhance endurance, safety, and features. Recently, planning and scheduling problems of drones have been studied in~\cite{UAV-communication}, \cite{UAV-refuel}. The authors in~\cite{UAV-communication} foresaw that drones could be used as mobile wireless communication nodes. In particular, the drones fly to collect data and return to transmit the data to the base station. The authors formulated the problem based on the pick-up and delivery VRP and added the communication constraints, which were not considered in the previous works. The authors in~\cite{UAV-refuel} addressed the drone path planning problem, which is also one type of VRP, by minimizing the total drone resource requirement, i.e., fuel or battery. In the problem, all the customers need to be visited once, and multiple depots are available for the drones to refuel or change their battery. The authors adopted resource constraint and developed a heuristic algorithm for the problem.
 
However, the authors in~\cite{UAV-communication} and \cite{UAV-refuel} did not consider uncertainty, which is unavoidable in a real environment. A stochastic drone mission planning is presented in~\cite{robust} and~\cite{online}. The authors in~\cite{robust} proposed a drone planning model with the stochastic resource consumption, i.e., fuel or energy consumption, traveling from one location to another location. The authors presented the comparison between the robust solutions and the solution with average resource consumption. Instead of considering stochastic resource consumption, the authors in~\cite{online} considered the uncertainty in recording time and traveling time from one location to another location, as well as a visiting time interval of each location, which is known as time-windows. The authors combined two objectives together by adding and balancing two weight parameters. The objectives are to maximize (i) the expected profits of visiting the locations which are already known before taking off and (ii) the percentage of reaching new locations, which are not known before taking off, on time. The new locations are emergency and unplanned cases. The authors proposed a re-planning approach as integer programming and developed a heuristic algorithm to solve it. However, in~\cite{robust} and \cite{online}, the authors formulated the problems as orienteering problems. In particular, a solution of the orienteering problem is to visit only some locations while the VRP requires all locations to be visited. 

For drone-related research, most of the existing works consider surveillance applications. Only few works are dedicated to parcel delivery~\cite{drone_delivery}. The drone parcel delivery can be found in~\cite{drone_delivery} and~\cite{sidekick}. The authors in~\cite{drone_delivery} proposed the drone delivery planning considering drones' capacity, battery weights, and changing payload weights. The authors modeled multiple drones, and each drone is able to serve multiple customers before returning to the depot. Drones are also allowed to come back to refuel or charge a battery at the depot for multiple times. The authors in~\cite{sidekick} considered the heterogeneous VRP for package delivery using both trucks and drones. The authors proposed the Flying Sidekick Traveling Salesman Problem (FSTSP) and the Parallel Drone Scheduling Traveling Salesman Problem (PDSTSP). In the FSTSP, the truck driver can set up a drone and let it visit different customers. After visiting and dropping packages to the assigned customers, the drone returns to the truck. Hence, the truck is referred to as a mobile depot for the drone separation and re-connection. On the other hand, in the PDSTSP, the authors formulated the problem with only one depot, but with multiple drones. Both the problems were formulated as a mix-integer programming model, where the objectives of the FSTSP and the PDSTSP are to minimize the serving time for both vehicles, i.e., one truck and one drone, and to minimize the traveling time of the truck, respectively. 

However, all the above works ignored the random breakdown of the drone delivery which can happen more frequently than land transportation, e.g., truck. Therefore, we propose the joint ground and aerial delivery service optimization and planning (GADOP) framework considering realistic drone operations. For example, (i) drones cannot take off due to weather conditions such as raining and strong wind, (ii) drones may have technical problems during flying, and (iii) drones can have an accident such as a bird strike. Therefore, we formulate an optimization of the GADOP framework as a stochastic integer programming problem. We model the uncertainty of drones including takeoff and breakdown conditions. Furthermore, to accelerate the computational time of solving the problem, we adopt L-shape~\cite{ref_L-shaped} to decompose the GADOP optimization. 

\section{System Model and Assumptions}
\label{sec_system}
\begin{figure}
\center\includegraphics[trim={0 12em 0 0},clip,width=0.5\textwidth]{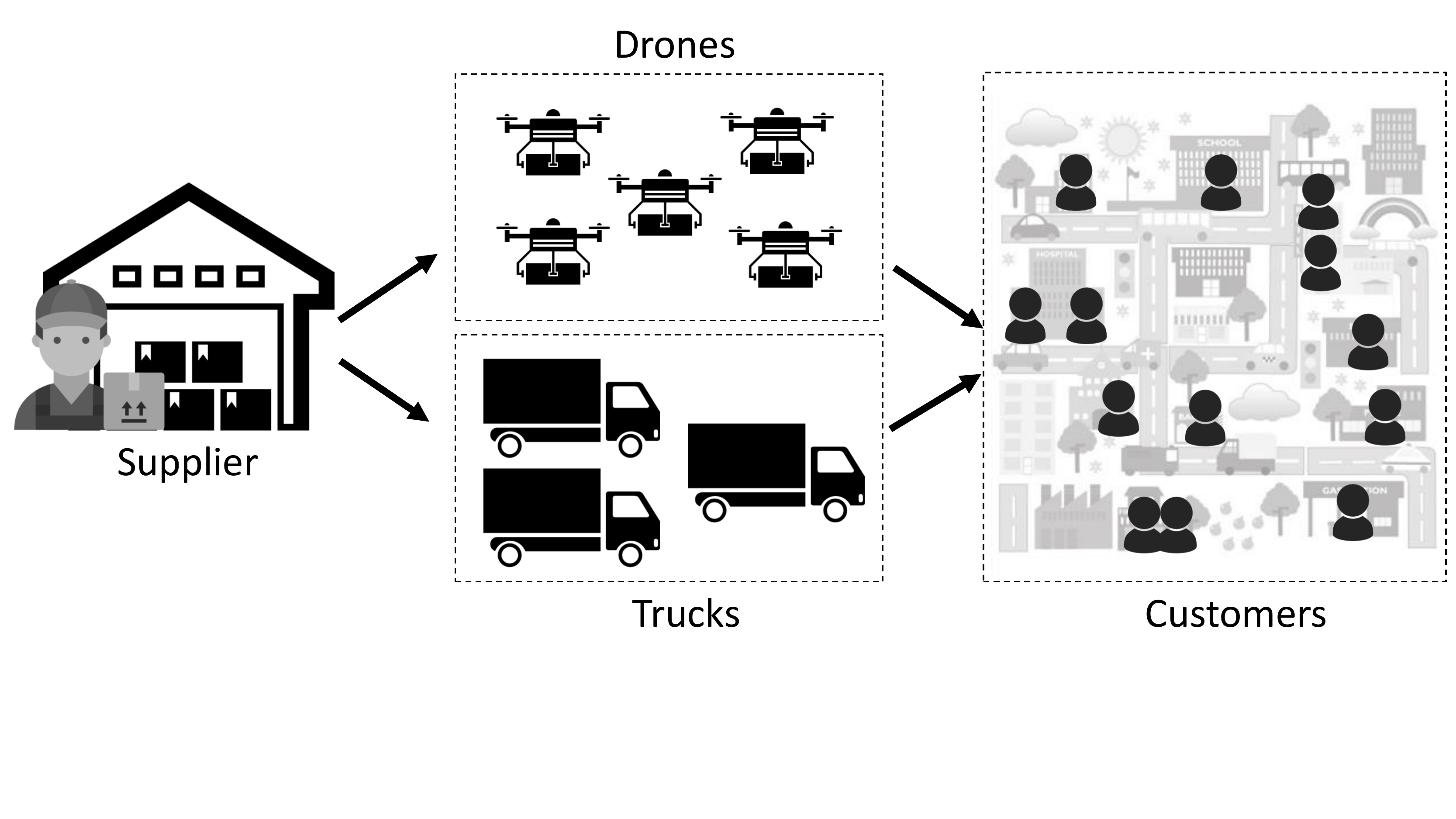} 
 \caption{The joint ground and aerial delivery service optimization and planning (GADOP) framework}
 \label{f_system}
\end{figure}

\begin{figure}
\center\includegraphics[width=0.5\textwidth]{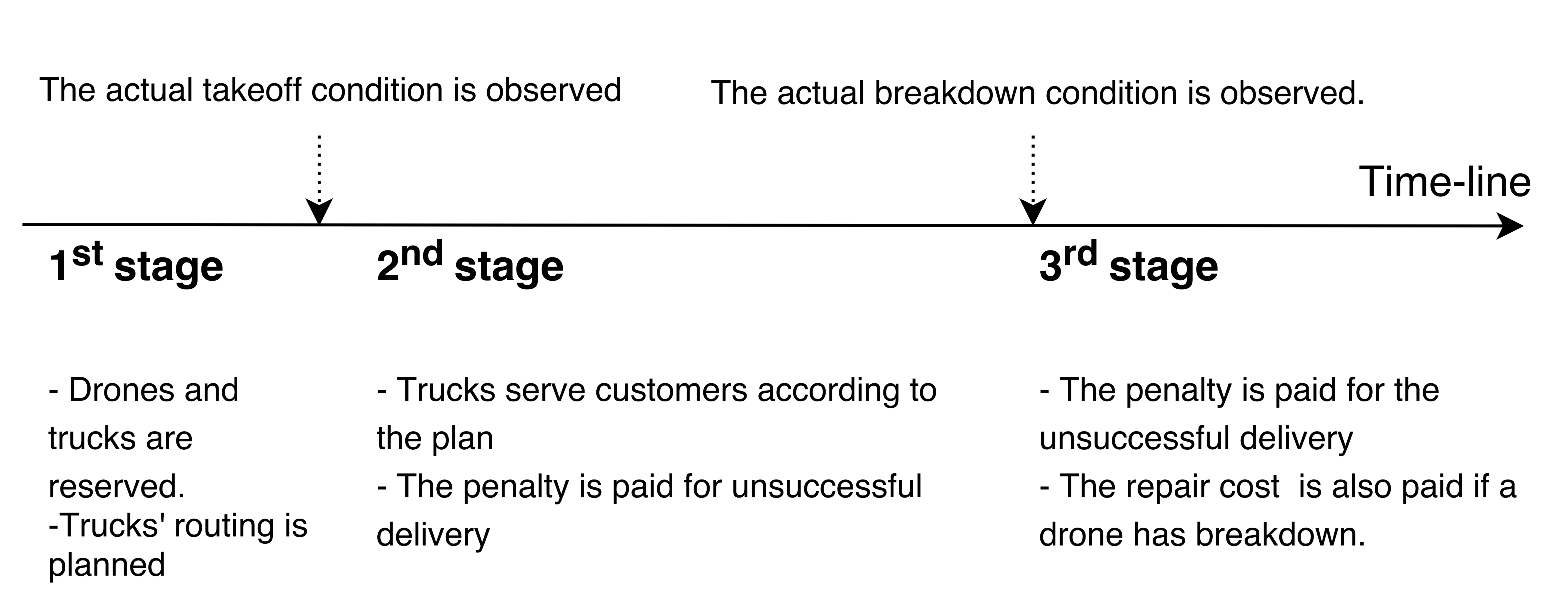} 
 \caption{The time-line of the scenarios.}
 \label{f_timeline}
\end{figure}

\begin{figure}
\center\includegraphics[width=0.5\textwidth]{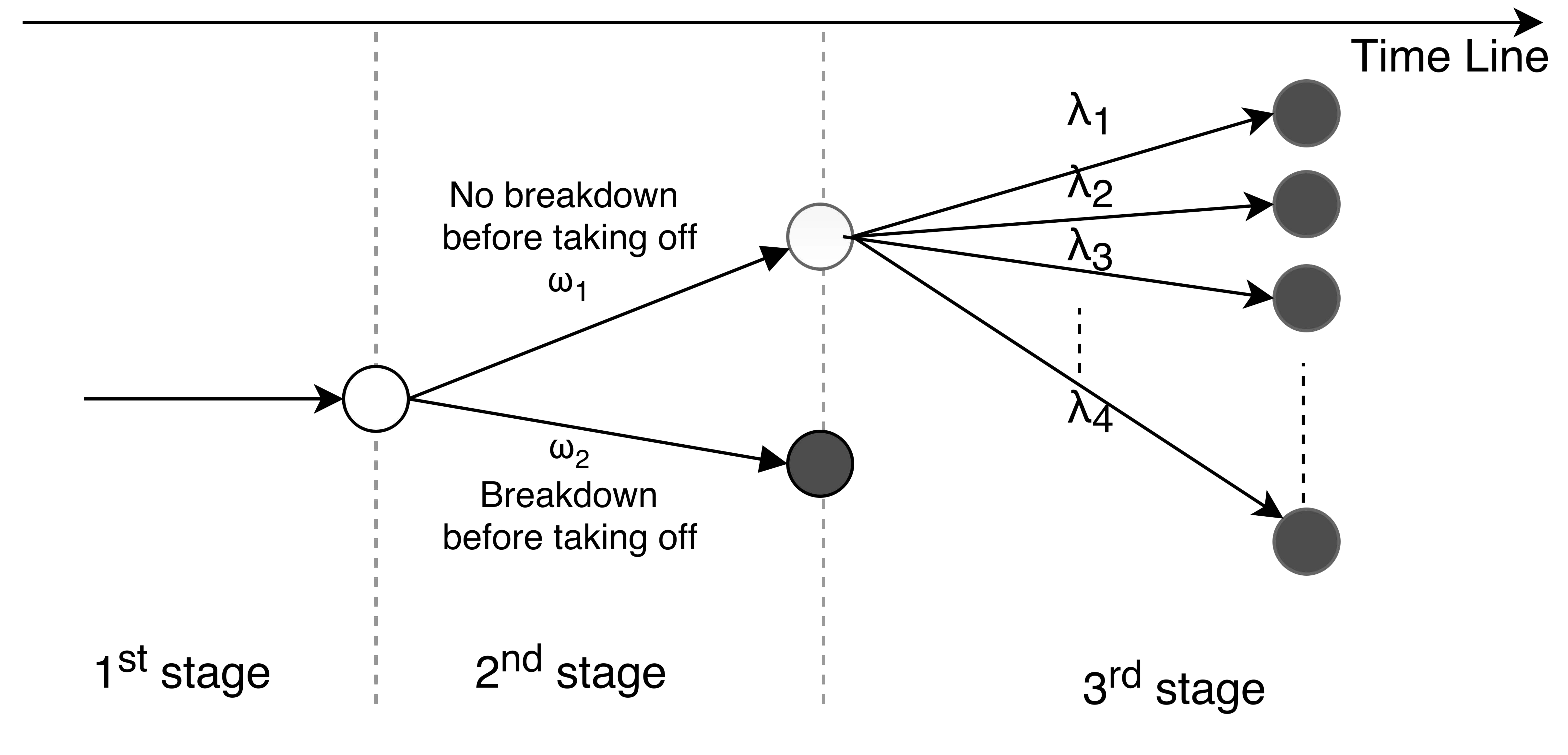} 
 \caption{An example of scenario tree with one drone. }
 \label{f_scenariotree}
\end{figure}

We consider package delivery service of a supplier. In this paper, we consider two delivery modes, i.e., truck delivery mode and drone delivery mode (Figure~\ref{f_system}). The truck delivery mode is for the supplier to rent vehicles, e.g., lorry trucks, and use the vehicles to deliver packages to customers. The drone delivery mode is for the supplier to rent drones and use the drones to deliver packages to customers. While trucks have a higher initial cost, e.g., rental fee and driver wage, and a higher traveling cost, i.e., fuel, than those of the drones, the trucks have larger capacity and are more reliable. 

Drone delivery is less reliable than the truck delivery as the drones can be affected by weather and other technical problems. Consequently, the drone delivery has a high breakdown probability. In this paper, we specifically consider two operability conditions of the drone, i.e., a takeoff condition and a breakdown condition. The takeoff condition indicates whether the drone can take off from the depot or not, which is observable before the drone departs. For example, the drone cannot takeoff from the depot if there is rain or strong wind. The breakdown condition happens after the drone already takes off from the depot, e.g., due to an accident. Therefore, the mulfunctioned drone needs to be collected from the site and repaired, which also incurs a certain cost. Furthermore, the supplier will pay a penalty composed of the fee for outsourcing the delivery to a carrier and the compensation to the customer due to the delayed delivery.

The supplier faces a problem of utilizing trucks and drones for package delivery for their customers. Trucks and drones need to be reserved in advance. Furthermore, the supplier has to take the uncertainties including drone takeoff and breakdown conditions into account. Therefore, we formulate the GADOP framework as a three-stage stochastic integer programming model. 
\begin{itemize}
\item \textbf{First Stage:} the supplier decides (i) whether customers will be served by either trucks or drones, (ii) how many trucks and drones need to be reserved, (iii) the routing path of trucks, and (iv) the serving order of drones' delivery. The decisions are made based on the probability of the uncertainties.
\item \textbf{Second Stage:} the supplier calculates the penalty payment for customers which cannot be served due to the drone takeoff condition. 
\item \textbf{Third Stage:} the supplier calculates the penalty payment for customers which cannot be served due to the drone breakdown condition. The breakdown condition is associated with each drone and each customer. Additionally, the supplier takes the repair cost into account when the breakdown happens. 
\end{itemize}
Figure~\ref{f_timeline} shows the stages of the GADOP framework. In the first stage, the decisions are made before the actual drone takeoff condition is observed. In the second stage, the decisions are made after the takeoff condition is observed and before the delivery begins. The third stage decisions are made when an actual breakdown of the drone is observed. Figure~\ref{f_scenariotree} shows an example of a scenario tree with one drone for the supplier. The formal definition of the scenarios is presented in Section~\ref{sec_uncertainty}. 

The supplier has a set of customers to be served denoted by $\mathcal{C}=\{C_1,C_2,\dots,C_{c'}\}$, where $c'$ denotes the total number of customers. Without loss of generality, each customer in $\mathcal{C}$ has one package to be delivered, and $g_i$ denotes the package weight of customer $i$. The supplier will assign customers' packages to either trucks or drones in the first stage. We assume that all the customers have a dropoff station for the drone delivery. The supplier can choose to reserve and use trucks from the set $\mathcal{T} = \{T_1,T_2, \dots,T_{t'}\}$ and drones from the set $\mathcal{D} = \{D_1,D_2, \dots,D_{d'}\}$. $t'$ and $d'$ denote the total numbers of available trucks and drones, respectively. Every truck in the set $\mathcal{T}$ has a capacity limit ($\bar{f}_t$), daily distance traveling limit ($\bar{l_t}$), and daily traveling time limit ($\bar{h_t}$). The package dropoff time of a truck is denoted by $\bar{r}$. Similarly, every drone in the set $\mathcal{D}$ has a capacity limit ($\widehat{f}_d$), daily flying distance limit ($\widehat{l}_d$), and flying distance limit per trip ($\widehat{e}_d$). The package dropoff time of a drone is assumed to be negligible. 

The supplier has one depot. The traveling distance from location $i'$ to location $j'$ is represented as $k_{i',j'}$, where $i'$ and $j'$ are the locations being members of the set $\mathcal{C}\cup\{0\}$, where $0$ represents the depot. The average truck driving speed and the drone flying speed are denoted by $\bar{q}_{i',j'}$ and $\widehat{q}_{i',j'}$, respectively. We assume that a drone can carry only one package per trip, and therefore the drone needs to return to the depot after delivering one package. A similar assumption is made in the literature, e.g.,~\cite{sidekick}, \cite{ref_roundtrip}.

\subsection{Uncertainty}
\label{sec_uncertainty}

The drone delivery is much less reliable than the truck delivery~\cite{ref_reliable}. Therefore, we consider the uncertainty of the drones including takeoff and breakdown conditions. Again, the takeoff condition and breakdown condition are observable in the second and third stages, i.e., before and after taking off, respectively. Let $\Omega=\{\omega_1,\omega_2, \dots,\omega_{\omega'}\}$ be the set of the takeoff scenarios, where $\omega'$ represents the total number of the scenarios. The takeoff scenario is defined as $\omega = (\mathbb{R}_1, \mathbb{R}_2, \dots, \mathbb{R}_{d'})$,  the maximum number of takeoff scenarios is based on the total number of drones, i.e., $d'$. $\mathbb{R}_d =1$ when drone $D_d$ cannot take off from the depot, and $\mathbb{R}_d =0$ otherwise. For example, $\omega = (0,0,1)$ means that the supplier has three drones, and drones $D_1$ and $D_2$ can take off while drone $D_3$ cannot take off from the depot. Let $\Lambda = \{\lambda_1,\lambda_2,\dots,\lambda_{\lambda'}\}$ denote the set of breakdown scenarios, where $\lambda_{\lambda'}$ represents the total number of the scenario. The breakdown  scenario $\lambda$ contains the breakdown parameters indicating which drones are broken during the delivery. The scenario is defined as follows: 
\begin{myalign}
\lambda = \begin{bmatrix}
\mathbb{B}_{1,1} & \mathbb{B}_{1,2} &\dots& \mathbb{B}_{1,d'} \\
\mathbb{B}_{2,1} & \mathbb{B}_{2,2} &\dots& \mathbb{B}_{2,d'} \\
 \vdots 		&  \vdots			&	\dots 	&\vdots\\
\mathbb{B}_{c',1} &\mathbb{B}_{c',2} & \dots&\mathbb{B}_{c',d'}
\end{bmatrix},
\label{eq_matrix}
\end{myalign}
where $\mathbb{B}_{i,d} =1$ when drone $D_d$ breaks down while serving customer $C_i$, and $\mathbb{B}_{i,d} = 0$ otherwise. Additionally, $\mathbb{P}(\omega)$ and $\mathbb{P}(\lambda)$ denote the probabilities that scenario $\omega$ and scenario $\lambda$ will happen, respectively. In reality, the probabilities can be obtained from the history records.

\subsection{Payments}
\label{sec_sys_payment}

The supplier involves six types of payments, i.e., costs, including (i) the initial costs of trucks, (ii) the initial costs of drones, (iii) the traveling costs of trucks, (iv) the traveling costs of drones, (v) penalty of unsuccessful drone delivery, and (vi) drone repair cost. The initial costs are the fixed costs such as rental fee, driver or staff wages. $\bar{c}_t^{(i)}$ and $\widehat{c}_d^{(i)}$ represent the initial costs of trucks and the initial costs of drones, respectively. The traveling costs are functions of traveling distance, fuel cost, or battery recharge cost. $\bar{c}_t^{(r)}$ and $\widehat{c}_d^{(r)}$ represent the traveling costs of trucks and the traveling costs of drones, respectively. The penalty and the repair cost, denoted as $p$ and $m$, respectively, are incurred when a drone does not successfully deliver a package. 


\section{Problem Formulation}
\label{sec_formulation}

\begin{align}
& \mbox{Minimize:} \nonumber \\
& \sum_{t \in \mathcal{T}}\bar{c}^{(i)}_t \bar{W}_{t} + 
\sum_{d \in \mathcal{D}}\widehat{c}^{(i)}_d \widehat{W}_d + \sum_{\substack{i', j'\in \\ \mathcal{C}\cup\{0\}}}\sum_{t\in \mathcal{T}} \bar{c}^{(r)}_{i',j'}V_{i',j',t} + \mathbb{E}(\mathscr{L}(\widehat{X}_{i,d})),
\label{eq_obj1}
\end{align}

\vspace{-3em}
\begin{align}
&\text{where}\nonumber \\
&\mathscr{L}(\widehat{X}_{i,d}) &=& \sum_{\omega \in \Omega} \mathbb{P}(\omega)
\left(
\sum_{i \in \mathcal{C}}\sum_{d \in \mathcal{D}}\widehat{c}^{(r)}_{i}\widehat{X}_{i,d}(1-\mathbb{R}_d(\omega) )+\right.\nonumber \\
&&&\left.\sum_{d \in \mathcal{D}} pZ^{(b)}_{i,d}(\omega) + \mathbb{E}(\mathscr{M}(\widehat{X}_{i,d}, \omega)) \right), \label{eq_obj2}\\
&\mathscr{M}(\widehat{X}_{i,d},\omega) &=& \sum_{\lambda \in \Lambda}\mathbb{P}(\lambda) \left(\sum_{i \in \mathcal{C} }\sum_{d \in \mathcal{D} }pZ^{(a)}_{i,d}(\omega,\lambda) \right.
\nonumber \\ 
&&&\left. + \sum_{d \in \mathcal{D}} mZ^{(m)}_{d}(\omega,\lambda)\right),
\label{eq_obj3}
\end{align}
\noindent subject to: (\ref{con_ini_ftl}) to (\ref{eq_con_lastbound}).

The GADOP framework is a three-stage stochastic integer programming as modeled in (\ref{eq_obj1}) for the first stage, (\ref{eq_obj2}) for the second stage, and (\ref{eq_obj3}) for the third stage. $\mathbb{E}(\mathscr{L}(\widehat{X}_{i,d}))$ denotes the expectation of the traveling cost and the penalty of drones over random takeoff conditions occurred in the second stage. $\mathbb{E}(\mathscr{M}(\widehat{X}_{i,d},\omega))$ denotes the expectation of the penalty cost and the repair cost of drones over random takeoff conditions and random breakdown conditions occurred in the third stage. The objective is to minimize the total delivery payments as discussed in Section~\ref{sec_sys_payment}. The decision variables in the framework are listed below.
\begin{itemize}
\item $\bar{W}_{t}$ is an indicator whether truck $t$ will be used or not. When $\bar{W}_{t} =1$, truck $t$ will be used, and $\bar{W}_{t}=0$ otherwise. 
\item $\widehat{W}_d$ is an indicator whether drone $d$ will be used or not. When $\widehat{W}_d =1$, drone $d$ will be used, and $\widehat{W}_d =0$ otherwise. 
\item $V_{i',j',t}$ is a routing variable in which $V_{i',j',t} =1$ if the route from location $i'$ to location $j'$ will be used by truck $t$, and $V_{i',j',t}=0$ otherwise. 
\item $\widehat{X}_{i,d}$ is an allocation variable of drone $d$. $\widehat{X}_{i,d}=1$ if drone $d$ will serve customer $i$, and $\widehat{X}_{i,d}=0$ otherwise. 
\item $\bar{X}_{i,t}$ is an allocation variable of truck $t$. $\bar{X}_{i,t} =1$ if truck $t$ will serve customer $i$, and $\bar{X}_{i,t}=0$ otherwise. 
\item $Z^{(b)}_{i,d}(\omega)$ is a penalty variable. $Z^{(b)}_{i,d}(\omega)=1$ if the penalty associated to customer $i$, which will be served by drone $d$, but the drone cannot take off from the depot, is paid in scenario $\omega$, and $Z^{(b)}_{i,d}(\omega)=0$ otherwise. 
\item $Z^{(a)}_{i,d}(\omega,\lambda)$ is another penalty variable. $Z^{(a)}_{i,d}(\omega,\lambda)=1$ if the penalty associated to customer $i$, which will be served by drone $d$, but the drone breaks down, is paid in scenarios $\omega$ and $\lambda$, $Z^{(a)}_{i,d}(\omega,\lambda)=0$ otherwise. 
\item $Z^{(m)}_{d}(\omega,\lambda)$ is a repair variable. $Z^{(m)}_{d}(\omega,\lambda)=1$ if drone $d$ in scenario $\omega$ and $\lambda$ breaks down during delivery, and $Z^{(m)}_{d}(\omega,\lambda)=0$ otherwise. 
\item $S_{i,t}$ is an auxiliary variable for sub-tour elimination in truck delivery.
\item $U_{i,d}$ is a serving order of the drones. $U_{i,d} < U_{j,d}$ means that drone $d$ will serve customer~$i$ before customer~$j$. 
\item $M_{i,j,d}$ is an auxiliary binary variable for ensuring the nonequivalent of  $U_{i,d}$. 
\end{itemize}
The types and bounds of all decision variables are indicated in constraints (\ref{eq_con_firstbound})-(\ref{eq_con_lastbound}).

The constraints in (\ref{con_ini_ftl}) and (\ref{con_ini_uav}) ensure that the initial cost of trucks and the initial cost of drones will be paid if they are used, where $\Delta$ is any large number $\Delta \geq c'$. The constraints in (\ref{con_capacity_ftl}) and (\ref{con_capacity_uav}) ensure that the total weight of packages does not exceed the capacity limit of a truck and a drone, respectively. The constraint in (\ref{con_fly_trip}) ensures that the flying distance between the depot to a customer and vice verse does not exceed the flying limit. The daily traveling distance limits are controlled by the constraints in (\ref{con_fly_day}) and (\ref{con_drive_day}) for drone delivery and truck delivery, respectively. The total time of the truck delivery, i.e., traveling time and serving time, must not exceed the working hours as imposed in the constraint in (\ref{con_serving}).

\begin{myalign}
&\sum_{i \in \mathcal{C}} \bar{X}_{i,t} \leq \Delta \bar{W}_{t}, & \forall t \in \mathcal{T} \label{con_ini_ftl}\\
&\sum_{i \in \mathcal{C}} \widehat{X}_{i,d} \leq \Delta \widehat{W}_d, & \forall d \in \mathcal{D}\label{con_ini_uav}\\
&\sum_{i\in \mathcal{C}}g_i\bar{X}_{i,t} \leq \bar{f}_t, &\forall t \in \mathcal{T} \label{con_capacity_ftl}\\
&g_i\widehat{X}_{i,d} \leq \widehat{f}_d, & \forall i \in \mathcal{C}, \forall d \in \mathcal{D}\label{con_capacity_uav}\\
&(k_{0,i}+k_{i,0})\widehat{X}_{i,d} \leq \widehat{e}_d, & \forall i \in \mathcal{C}, \forall d \in \mathcal{D} \label{con_fly_trip}\\
&\sum_{i \in C}(k_{0,i}+k_{i,0})\widehat{X}_{i,d} \leq \widehat{l}_d, & \forall d \in \mathcal{D} \label{con_fly_day}\\
&\sum_{\substack{i', j'\in \\ \mathcal{C}\cup\{0\}}}k_{i',j'}V_{i',j',t} \leq \bar{l}_{t}, & \forall t \in \mathcal{T} \label{con_drive_day}
\end{myalign}

The allocation of trucks and drones is imposed by the constraint in (\ref{con_allocation}), where one customer can be served by either a truck or a drone. The constraints in (\ref{con_routing_1})-(\ref{con_subtour}) are to find the route for truck delivery. Only the allocated truck can take the arrival and departing route to serve the customer. The constraint in (\ref{con_subtour}) ensures that no subtour exists in the solution. 

If drones cannot take off from the depot, the penalty must be paid for all associated customers, as indicated in the constraint in (\ref{con_2_stage}). In the case that the drones can take off, but they are broken after departing, the constraint in (\ref{con_3_stage}) ensures that the penalty must be paid. When breakdown occurs after departing from the depot, the drone will be no longer able to serve all the next customers. Consequently, the penalty associated to these customers must be paid as indicated in the constraint in (\ref{con_order_main}). Note that the next customer $j$ after customer $i$ has a higher serving order, i.e., $U_{j,d} > U_{i,d}$. Moreover, the constraint in (\ref{con_repair}) ensures that the repair cost must be paid if the drone is broken after departing from the depot. 

The constraints in (\ref{eq_con_order3})-(\ref{eq_con_order2}) decide serving orders of drones. The constraint in (\ref{eq_con_order3}) ensures that $U_{i,d}$ must not be zero when  drone $d$ is allocated to serve customer $i$, i.e., $\ddot{X}_{i,d}=1$. $U_{i,d} = 0$ which represents the drone serving order is thus not needed. The constraints in (\ref{eq_con_order1})-(\ref{eq_con_order2}) ensure that the serving orders must not be the same when $\ddot{X}_{i,d}=1$. Again, $M_{i,j,d}$ is an auxiliary binary variable for ensuring that  $U_{i,d} \neq U_{j,d}$ when $\widehat{X}_{i,d} =1$.

\begin{myalign}
&\sum_{\substack{i', j'\in \\ \mathcal{C}\cup\{0\}}} \left( \frac{k_{i',j'}}{q_{i',j'}}+ \bar{r} \right) V_{i',j',t} \leq \bar{h}_t, & \forall t \in \mathcal{T}
\label{con_serving}\\
&\sum_{t\in \mathcal{T}}\bar{X}_{i,t} + \sum_{d \in \mathcal{D}} \widehat{X}_{i,d} = 1, & \forall i \in \mathcal{C} \label{con_allocation}
\end{myalign}

\begin{myalign}
&\sum_{i \in \mathcal{C}} V_{0,i,t}\leq 1, & \forall t \in \mathcal{T} \label{con_routing_1}\\
&\sum_{i \in \mathcal{C}} V_{i,0,t} \leq 1, & \forall t \in \mathcal{T}\\
&\sum_{i' \in \mathcal{C}\cup\{0\}} V_{i',i,t} = \bar{X}_{i,t}, & \forall t \in \mathcal{T}, \forall i \in \mathcal{C}\\
&\sum_{i' \in \mathcal{C}\cup\{0\}} V_{i,i',t} = \bar{X}_{i,t}, & \forall t \in \mathcal{T}, \forall i \in \mathcal{C}\\
& V_{i',i',t} = 0, & \forall t \in \mathcal{T}, \forall i' \in \mathcal{C}\cup\{0\}
\end{myalign}

\begin{myalign}
&S_{i,t} - S_{j,t} + c'V_{i,j,t} \leq c'-1,\hspace{0.5em} \forall i \in \mathcal{C}, \forall j \in \mathcal{C}, \forall t \in \mathcal{T} \label{con_subtour} \\
&\widehat{X}_{i,d}\mathbb{R}_d(\omega) = Z^{(b)}_{i,d}(\omega), \hspace{2.8em} \forall i \in \mathcal{C}, \forall d \in \mathcal{D}, \forall \omega \in \Omega\label{con_2_stage}\\ \nonumber
&\widehat{X}_{i,d}(1-\mathbb{R}_d(\omega))\mathbb{B}_{i,d}(\lambda) \leq Z^{(a)}_{i,d}(\omega,\lambda), \\ 
& \hspace{9em} \forall i \in \mathcal{C}, \forall d \in \mathcal{D}, \forall \omega \in \Omega, \lambda \in \Lambda \label{con_3_stage}\\ \nonumber
&\left( U_{i,d}- U_{j,d}\right) \leq \Delta \left( 1 - Z^{(a)}_{j,d}(\omega,\lambda) + Z^{(a)}_{i,d}(\omega,\lambda) \right),\\
& \hspace{5.5em}\forall i, j \in \mathcal{C}, i\neq j, \forall d \in \mathcal{D}, \forall \omega \in \Omega, \lambda \in \Lambda\label{con_order_main}\\ \nonumber 
&Z^{(m)}_{d}(\omega,\lambda) \geq Z^{(a)}_{i,d}(\omega,\lambda), \\ 
& \hspace{9em}\forall i \in \mathcal{C}, \forall d \in \mathcal{D}, \forall \omega \in \Omega, \lambda \in \Lambda\label{con_repair}
\end{myalign}
\begin{myalign}
&\widehat{X}_{i,d} \leq U_{i,d}, & \forall d \in \mathcal{D}, \forall i \in \mathcal{C}\label{eq_con_order3}\\
&0 \leq U_{j,d} \leq \sum_{i \in \mathcal{C}}\widehat{X}_{i,d}, & \forall d \in \mathcal{D}, \forall j \in \mathcal{C}\label{eq_con_order4}
\end{myalign}
\begin{myalign}
&U_{i,d} - U_{j,d} \leq \Delta M_{i,j,d}-\widehat{X}_{i,d}, \nonumber\\
& \hspace{12em}\forall i ,j \in \mathcal{C}, i \neq j,\forall d \in \mathcal{D}
 \label{eq_con_order1}\\ 
&U_{i,d} - U_{j,d} \geq \widehat{X}_{i,d} - \Delta( 1- M_{i,j,d}), \nonumber\\
& \hspace{12em}\forall i ,j \in \mathcal{C}, i \neq j,\forall d \in \mathcal{D}
 \label{eq_con_order2}
\end{myalign}

In order to enhance customers' satisfaction, suppliers allow their customers to specific a delivery time slot, which can be referred to as time windows. Time windows constraint can be hard  time windows or soft time windows.  In the hard time window, all customers must be served within time windows. On the other hand, the soft time window is more relaxed, and thus the supplier can choose to pay for the penalty of not delivery on time. We consider the hard time window in this paper. The constraints in (\ref{eq_con_tw_1})-(\ref{eq_con_tw_2}) ensures that the delivery is done within the assigned time windows.  $\mathcal{L}^{(m)}$ and $\mathcal{L}^{(a)}$ denote a set of customers who require to be served in the morning and afternoon, respectively, and we have $\mathcal{L}^{(m)} \subseteq \mathcal{C}$, $\mathcal{L}^{(a)} \subseteq \mathcal{C}$ and $\mathcal{L}^{(m)} \cap \mathcal{L}^{(a)} = \emptyset$. The constraints in (\ref{eq_con_tw_1}) and (\ref{eq_con_tw_2}) ensure that all morning customers must be served before the afternoon customers. The constraints in (\ref{eq_con_tw_d_1})-(\ref{eq_con_tw_t_2}) ensure that traveling time of trucks and drones does not exceed the morning and afternoon traveling time limits, i.e., time windows. The morning and afternoon traveling time limits  are defined as $l^{(m)}$ and $l^{(a)}$, respectively. More specific time slots and explaination of hard time window can be found in~\cite{ref_VTC1}.

\begin{myalign}
& U_{i,d} \leq U_{j,d} + \Delta(1- \widehat{X}_{i,t}), \nonumber
\\& \qquad\qquad\qquad\forall i \in \mathcal{L}^{(m)}, \forall j \in \mathcal{L}^{(a)}, \forall d \in \mathcal{D}, \forall \omega \in \Omega  \label{eq_con_tw_1}\\
& S_{i,t} \leq S_{j,t} + \Delta(1- \bar{X}_{i,t}),\nonumber\\ 
&\qquad\qquad\qquad\forall i \in \mathcal{L}^{(m)}, \forall j \in \mathcal{L}^{(a)}, \forall t \in \mathcal{T}, \forall \omega \in \Omega  \label{eq_con_tw_2}
\end{myalign}

\begin{myalign}
& \sum_{m \in \mathcal{L}^{(m)}}\left(\frac{k_{0,m}}{\widehat{q}_{0,m}} + \frac{k_{m,0}}{\widehat{q}_{m,0}}\right)\widehat{X}_{m,d} \leq l^{(m)},& \forall d \in \mathcal{D} \label{eq_con_tw_d_1}\\
&  \sum_{a \in \mathcal{L}^{(a)}}\left(\frac{k_{0,a}}{\widehat{q}_{0,a}} + \frac{k_{a,0}}{\widehat{q}_{a,0}}\right)\widehat{X}_{a,d} \leq l^{(a)},& \forall d \in \mathcal{D} \label{eq_con_tw_d_2}\\
&\sum_{\substack{i'\in \\ \mathcal{C}\cup\{0\}}}\sum_{\substack{ m \in \mathcal{L}^{(m)}}} \left( \frac{k_{i',m}}{\bar{q}_{i',m}}+ \bar{r} \right) V_{i',m,t} \leq l^{(m)}, & \forall t \in \mathcal{T} \label{eq_con_tw_t_1}\\
&\sum_{\substack{i\in  \mathcal{C}}}\sum_{\substack{ a' \in\\ \mathcal{L}^{(a)}\cup\{0\}}} \left( \frac{k_{i,a'}}{\bar{q}_{i,a'}}+ \bar{r} \right) V_{i,a',t} \leq l^{(a)}, & \forall t \in \mathcal{T} \label{eq_con_tw_t_2}
\end{myalign}

\begin{myalign}
& \widehat{X}_{i,d}, \bar{X}_{i,t}, \widehat{W}_d, \bar{W}_t \in \{0,1\},\hspace{1em} \forall d \in \mathcal{D}, \forall t \in \mathcal{T}, \forall i \in \mathcal{C} \label{eq_con_firstbound}\\
& V_{i',j',t} \in \{0,1\}, \hspace{5em} \forall i ',j' \in \mathcal{C}\cup\{0\}, \forall t \in \mathcal{T}\\
& U_{i,d}, S_{i,t} \in \{0,1,\dots, c'\}, \hspace{1.5em} \forall d \in \mathcal{D},  \forall t \in \mathcal{T}, \forall i \in \mathcal{C} \label{eq_con_secondlast}\\
& Z^{(b)}_{i,d}(\omega), Z^{(a)}_{i,d}(\omega,\lambda), Z^{(m)}_{d}(\omega,\lambda), M_{i,j,d} \in \{0,1\}, \nonumber \\
&\hspace{9em} \forall d \in \mathcal{D}, \forall i,j \in \mathcal{C}, \omega \in \Omega, \lambda \in \Lambda
\label{eq_con_lastbound}
\end{myalign}

The above optimization problem can be complex to solve when the numbers of customers, trucks, and drones grow. Therefore, in the next section, we present the decomposition method to obtain an optimal solution of the problem more efficiently.


\section{Decomposition}
\label{sec_decom}

The GADOP optimization, which is presented in the last section, is an NP-hard problem. The computational time for solving the problem depends on the numbers of variables, parameters, and constraints. To address the computational issue, we decompose the GADOP optimization into multiple sub-problems according to the decision stages, i.e., one first-stage problem, $\omega'$ second-stage sub-problems, and $\omega'$ third-stage sub-problems. Each second-stage sub-problem and each third-stage sub-problem are associated with a certain takeoff scenario, i.e., $\omega \in \Omega$. The decomposition is executed iteratively as described in the following steps:
\begin{enumerate}[(i)]
\item Solving the first-stage problem, which is the master problem, 
\item Solving all the second-stage sub-problems and the third-stage sub-problems using the results from the master problem as the parameters, 
\item Calculating feedback parameters, i.e., $E^{(p)}_{i,d}$, $E^{(r)}_{i,d}$, and $E^{(m)}_{i,d}$ representing the penalty, the traveling cost of drones, and drone repair cost from the second-stage sub-problems and third-stage sub-problem, and 
\item Checking the convergence condition, if the convergence condition is met, then the solution is found, and otherwise the feedback parameters are sent to the master problem, and Step (i) repeats with the additional constraints applying the feedback parameters.
\end{enumerate}
This decomposition is based on the L-Shape method~\cite{ref_L-shaped}. The algorithm to execute the decomposition is shown in Algorithm~\ref{al_decom}.

\begin{figure}
\center
\includegraphics[width=0.5\textwidth]{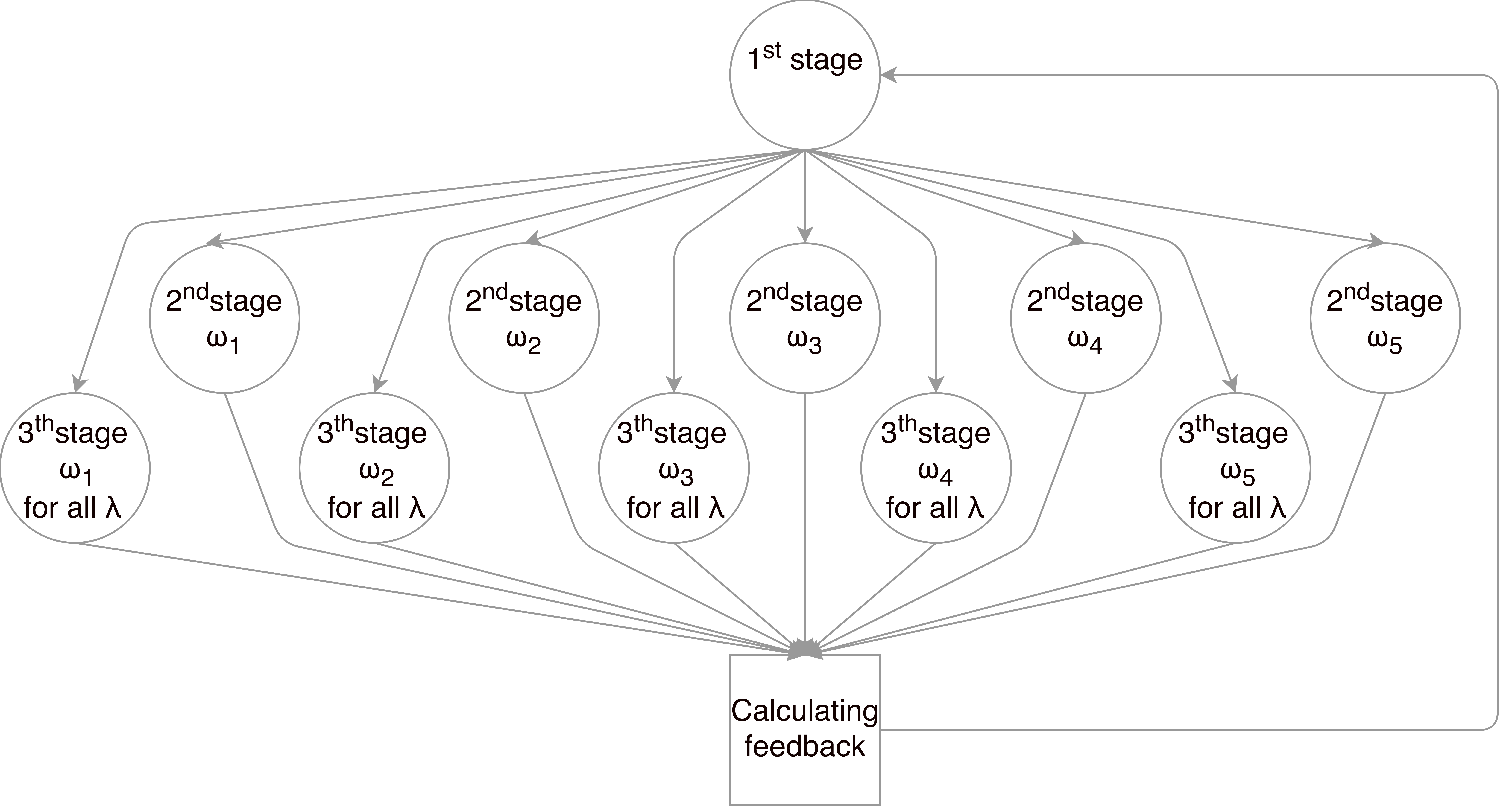} 
 \caption{Decomposition of scenarios and stages where a circle represents one optimization master problem or sub-problem.}
 \label{f_parallel}
\end{figure}

Figure~\ref{f_parallel} presents an example of the GADOP optimization with five takeoff scenarios and ten breakdown scenarios. After decomposed, the example has one master problem, five second-stage sub-problems, and five third-stage sub-problems to be solved. Note that the numbers of the second-stage sub-problems and the third-stage sub-problems depend on the number of the takeoff scenarios. In fact, after applying the decomposition, we have found that the solution of the second-stage problem can be obtained by a simple calculation due to special structure of the GADOP problem. As a result, for the same takeoff scenario, the second-stage sub-problem and the third-stage sub-problem do not have any interdependency with each other, and thus we can solve the second-stage sub-problems and the third-stage sub-problems simultaneously. Additionally, a takeoff scenario does not have any interdependency with other takeoff scenarios. However, dependencies of breakdown scenarios exist in serving orders ($U_{i,d}$). Since the serving orders need to be decided before the delivery begins, the same serving order ($U_{i,d}$) is used for all breakdown scenarios. As a result, the third-stage optimization cannot be further decomposed according to breakdown scenarios.

\subsection{Master Problem or First-Stage Problem}
\noindent Minimizing:
\begin{myalign} 
\sum_{t \in \mathcal{T}}\bar{c}^{(i)}_t \bar{W}_{t} + 
\sum_{d \in \mathcal{D}}\widehat{c}^{(i)}_d \widehat{W}_d + \sum_{\substack{i', j'\in \\ C\cup\{0\}}}\sum_{t\in \mathcal{T}} \bar{c}^{(r)}_{i',j'}V_{i',j',t} + \theta_1 + \theta_2,
\label{eq_obj_dec_master}
\end{myalign}
subject to: (\ref{con_ini_ftl}) to (\ref{con_subtour}), (\ref{eq_con_tw_2}) to (\ref{eq_con_tw_t_2}), (\ref{eq_cutting_theta1}), and (\ref{eq_cutting_theta2}).
\begin{myalign}
&\sum_{i \in \mathcal{C}}\sum_{d \in \mathcal{D}}(E^{(p)}_{i,d} + E^{(r)}_{i,d} )\widehat{X}_{i,d} + \theta_1 \geq 0 &
\label{eq_cutting_theta1}\\
&\sum_{d \in \mathcal{D}}E^{(m)}_{i,d}\widehat{W}_{d} + \theta_2 \geq 0, & \forall i \in \mathcal{C} 
\label{eq_cutting_theta2}
\end{myalign}

The objective function of the master problem is presented in~(\ref{eq_obj_dec_master}). The constraints in (\ref{eq_cutting_theta1}) and (\ref{eq_cutting_theta2}) are the feedback cutting constraints~\cite{ref_L-shaped}. Again, $E^{(p)}_{i,d}$, $E^{(r)}_{i,d}$, and $E^{(m)}_{i,d}$ are the feedback parameters, which are dedicated to the penalty, the drone traveling cost, and the drone repair cost, respectively. $\theta_1$ and $\theta_2$ are initially set at small values for the first iteration. After the first iteration, $\theta_1$ and $\theta_2$ become positive decision variables. The results of $\theta_1$ and $\theta_2$ are denoted by $\theta^*_1$ and $\theta^*_2$, respectively, which are used in checking the convergence condition. The result of $\widehat{X}_{i,d}$ is denoted by $\widehat{X}^{*}_{i,d}$, which is used as a parameter in the second-stage sub-problems and the third-stage sub-problems. After decomposition, the first-stage problem is still an NP-hard problem. However, the first-stage problem becomes less complex than the original GADOP because the number of decision variables in the first-stage problem is fewer than that in the original problem.

\subsection{Second-Stage Sub-Problem }
\noindent Minimize:
\begin{myalign}
& \sum_{i \in \mathcal{C}}\sum_{d \in \mathcal{D}}\widehat{c}^{(r)}_{i}\widehat{X}^{*}_{i,d}(1-\widetilde{\mathbb{R}}_d )+
\sum_{d \in \mathcal{D}} p\widetilde{Z}^{(b)}_{i,d} ,
\label{eq_obj_dec_2}
d\end{myalign}
subject to: (\ref{eq_dec_2}).
\begin{myalign}
&\widehat{X}^{*}_{i,d}\widetilde{\mathbb{R}}_d = \widetilde{Z}^{(b)}_{i,d}, & \forall i \in \mathcal{C}, \forall d \in \mathcal{D} 
\label{eq_dec_2}
\end{myalign}

We decompose the second-stage sub-problem given a certain takeoff scenario, and thus the variable $\omega$ is removed from the objective function in (\ref{eq_obj_dec_2}) and the constraint in (\ref{eq_dec_2}). To minimize the expected cost in the second stage, i.e., $\mathbb{E}(\mathscr{L}(\widehat{X}_{i,u})) $, we need to solve all the second-stage sub-problems associated to a takeoff scenario ($\omega$). Note that random parameters $\widetilde{\mathbb{R}}_d$ and $\widetilde{\mathbb{B}}_{i,d}$, i.e., for scenario variables ${\mathbb{R}}_d$ and ${\mathbb{B}}_{i,d}$ defined in Section~\ref{sec_uncertainty}, respectively, as well as the decision variable $\widetilde{Z}^{(b)}_{i,d}$ is associated with the takeoff scenario, and therefore the $\omega$ is omitted.

In general, the second-stage problem is an optimization, and thus we present the second-stage problem in the optimization form. Nevertheless, in our exact problem, there is a special structure that makes the second-stage problem is able to be solved directly through the assignment of $\widehat{X}_{i,d}$ and $\widetilde{\mathbb{R}}_d$.

After applying the decomposition, there is only one decision variable in the second-stage problem that is $\widetilde{Z}^{(b)}_{i,d}$. Furthermore, the decision variable $\widehat{X}_{i,d}$ is solved in the first-stage problem, and thus it becomes a parameter, i.e., $\widehat{X}^*_{i,d}$, in the second-stage problem. According to the constraint in (\ref{eq_dec_2}), $\widetilde{Z}^{(b)}_{i,d}$ depends on parameters only, and consequently the second-stage problem is no longer an optimization problem.


\subsection{Third-Stage Sub-Problem}
\noindent Minimize:
\begin{myalign}
& \sum_{\lambda \in \Lambda}\mathbb{P}(\lambda)\left(\sum_{i \in \mathcal{C} }\sum_{d \in \mathcal{D} }p\widetilde{Z}^{(a)}_{i,d}(\lambda)
+ \sum_{d \in \mathcal{D}}m\widetilde{Z}^{(m)}_{d}(\lambda)\right),
\label{eq_obj_dec_3}
\end{myalign}
subject to: (\ref{eq_con_order3}), (\ref{eq_dec_3_1}) to (\ref{eq_dec_order2}).
\begin{myalign}
&\widehat{X}^{*}_{i,d}(1-\widetilde{\mathbb{R}}_d)\widetilde{\mathbb{B}}_{i,d} \leq \widetilde{Z}^{(a)}_{i,d}(\lambda), \hspace{3.5em} \forall i \in \mathcal{C}, \forall d \in \mathcal{D} \label{eq_dec_3_1}\\\nonumber
& U_{i,d}- U_{j,d} \leq \Delta \left( 1 - \widetilde{Z}^{(a)}_{j,d}(\lambda) + \widetilde{Z}^{(a)}_{i,d}(\lambda) \right), \\ 
& \hspace{12.5em} \forall i, j \in \mathcal{C}, i\neq j, \forall d \in \mathcal{D} \label{eq_dec_3_2}\\
&\widetilde{Z}^{(m)}_{d}(\lambda) \geq \widetilde{Z}^{(a)}_{i,d}(\lambda), \hspace{7.5em} \forall i \in \mathcal{C}, \forall d \in \mathcal{D}\label{eq_dec_3_3}\\
&\widehat{X}^{*}_{i,d} \leq U_{i,d},\hspace{10.5em} \forall d \in \mathcal{D}, \forall i \in \mathcal{C}\label{eq_dec_order1} \\
&0 \leq U_{j,d} \leq \sum_{i \in \mathcal{C}}\widehat{X}^{*}_{i,d}, \hspace{7em} \forall d \in \mathcal{D}, \forall j \in \mathcal{C}\label{eq_dec_order2}\\
&U_{i,d} - U_{j,d} \leq \Delta M_{i,j,d}-\widehat{X}^*_{i,d}, \nonumber\\
& \hspace{12em}\forall i ,j \in \mathcal{C}, i \neq j,\forall d \in \mathcal{D}
 \label{eq_dec_order3}\\ 
&U_{i,d} - U_{j,d} \geq \widehat{X}^*_{i,d} - \Delta( 1- M_{i,j,d}), \nonumber\\
& \hspace{12em}\forall i ,j \in \mathcal{C}, i \neq j,\forall d \in \mathcal{D}
 \label{eq_dec_order4}
\end{myalign}

Similar to the second-stage sub-problem, each third-stage sub-problem is formulated given a certain takeoff scenario. The objective function of the third-stage sub-problem is expressed in (\ref{eq_obj_dec_3}) which is the expected cost of the penalty and repair cost of drones. The constraints in (\ref{eq_dec_3_1}), (\ref{eq_dec_3_2}), and (\ref{eq_dec_3_3}) contain the third-stage decision variables, which are $\widetilde{Z}^{(a)}_{i,d}(\lambda)$ and $\widetilde{Z}^{(m)}_{d}(\lambda)$. The decision variables $\widetilde{Z}^{(a)}_{i,d}(\lambda)$ and $\widetilde{Z}^{(m)}_{d}(\lambda)$ are associated with the takeoff scenario, and therefore $\omega$ is omitted. Since there is the interdependency among $U_{i,d}$, i.e., drone serving order, the constraints in (\ref{eq_dec_order1})-(\ref{eq_dec_order4}) are directly adopted in this sub-problem. The reason is that $U_{i,d}$ of each takeoff scenario, which depends on $\widehat{X}^{*}_{i,d}$, has already obtained in the master problem. Again, $\widehat{X}^{*}_{i,d}$ is the assignment of a drone to a customer.

\subsection{Calculating Feedback Parameters and Convergence}

This section explains the feedback parameter calculation and the convergence condition checking. $e_{i,d}^{(p)}(\omega)$ and $e_{i,d}^{(m)}(\omega)$ are the auxiliary parameters in the third-stage sub-problems, which are associated with the drone penalty and the drone repair cost, respectively. The feedback parameters, i.e., $E_{i,d}^{(r)}$, $E_{i,d}^{(p)}$, and $E_{i,d}^{(m)}$,  as expressed in (\ref{eq_feedback_routing}), (\ref{eq_feedback_penalty}), and (\ref{eq_feedback_maintenance}),   respectively, are returned to the master problem. $E_{i,d}^{(r)}$, $E_{i,d}^{(p)}$, and $E_{i,d}^{(m)}$ are the expected values associated with the drone traveling cost, the drone penalty, and the drone repair cost, respectively. The constraints in (\ref{eq_cutting_theta1}) and (\ref{eq_cutting_theta2}) ensure that the supplier pays the routing cost, the penalty, and the repair cost from the second-stage and third-stage when the associated $\widehat{X}_{i,d}$ is selected. Therefore, $E_{i,d}^{(r)}$, $E_{i,d}^{(p)}$, and $E_{i,d}^{(m)}$ must be negative  in order to match with $\theta_1$ and $\theta_2$. The formulations of the feedback parameters are given as follow.

\begin{myalign}
E_{i,d}^{(r)} &&=& -\widehat{c}^{(r)}_{i,d}\sum_{\omega \in \Omega}\mathbb{P}(\omega)\left( 1- \mathbb{R}_d(\omega) \right),
\label{eq_feedback_routing}\\
E_{i,d}^{(p)} &&=& -\sum_{\omega \in \Omega}\mathbb{P}(\omega)\left( p\mathbb{R}_d(\omega) + e_{i,d}^{(p)}(\omega) \right),
\label{eq_feedback_penalty}\\
\text{where } e_{i,d}^{(p)}(\omega) &&=& p\sum_{\lambda \in \Lambda}\mathbb{P}(\lambda)\left(1-\mathbb{R}_d(\omega)\right)\mathbb{B}_{i,d}(\lambda),
\label{eq_feedback_penalty_third}\\
E_{i,d}^{(m)} &&=& -\sum_{\omega \in \Omega}\mathbb{P}(\omega) e_{i,d}^{(m)}(\omega),
\label{eq_feedback_maintenance}\\
\text{where }e_{i,d}^{(m)}(\omega) &&=& m\sum_{\lambda \in \Lambda}\mathbb{P}(\lambda)\left(1-\mathbb{R}_d(\omega)\right)\mathbb{B}_{i,d}(\lambda).
\label{eq_feedback_maintenance_third}
\end{myalign}

We next define a convergence parameter denoted by $B$ which is used for checking the convergence condition. If $B$ is less than or equal to $\theta_1^* + \theta_2^*$, the convergence condition is met, and then Algorithm~\ref{al_decom} is terminated and the solution is taken from the latest iteration. The convergence parameter is calculated from (\ref{eq_convergence_1}) and (\ref{eq_convergence_2}). 

The proof of convergence of the L-shape method can be found in~\cite{ref_L-shaped}. In the GADOP optimization problem, we observe that the feedback constraints are based on the probabilities of scenarios, the penalty, the routing cost, and the repair cost. As we observe from the GADOP problem, these feedback parameters can be calculated directly from parameters, i.e., the probabilities of  scenarios ($\mathbb{P}(\omega),\mathbb{P}(\lambda)$), the penalty ($p$), the routing cost ($\widehat{C}_{i,d}^{(r)}$), and the repair cost ($m$). Since the feedback parameters do not depend on the solution at an iteration $k$, the feedback terms are constants. 
According to Algorithm~1, we observe that the first-stage solutions at $k=1$ and $k=2$ are identical because the feedback terms are the same. Therefore, the convergence of the decomposed GADOP is guaranteed at $k=2$.

\begin{myalign}
&B = \sum_{i \in \mathcal{C} }\sum_{d \in \mathcal{D} }\widehat{X}^{*}_{i,d}(E_{i,d}^{(p)}+E_{i,d}^{(r)}) + M_d\widehat{W}^{*}_{d},\label{eq_convergence_1} \\
&M_d = \max(E_{i,d}^{(m)}), \hspace{7em}\forall i \in \mathcal{D}\label{eq_convergence_2}
\end{myalign}

\begin{algorithm}[t]

\fbox{\begin{minipage}[b]{0.45\textwidth} 
\SetAlgoLined

\KwResult{The optimal solution is taken from the latest iteration }

 \While{ Convergence condition is not met}{
 
 \eIf{$k=0$}
 {
 {\ttfamily //first iteration}\;
 - $\theta_1^{*k}$ and $\theta_2^{*k}$ are set at very small values\; 
 - Solving master problem without (\ref{eq_cutting_theta1}) and (\ref{eq_cutting_theta2})\;
 }{
 - Generating the feedback constraints as in (\ref{eq_cutting_theta1}) and (\ref{eq_cutting_theta2}) by using $E^{(p),k-1}_{i,d}$, $E^{(m),k-1}_{i,d}$, and $E^{(r),k-1}_{i,d}$ \;
 - Solving master problem with the feedback constraints\;
 }
 - Obtaining $X^{*k}_{i,d}$, $\theta_1^{*k}$, and $\theta_2^{*k}$\;

 \ForEach{ scenario $\omega \in \Omega$}{%
 - Solving the second-stage sub-problem by using $X_{i,d}^{*k}$ as parameters\;
 - Solving the third-stage sub-problem by using $X_{i,d}^{*k}$ as parameters\; 
 }
 
 - Calculating\\ 
 $E^{(p),k}_{i,d}$, $E^{(m),k}_{i,d}$, and $E^{(r),k}_{i,d}, \hspace{3em}\forall d \in \mathcal{D}, \forall i \in \mathcal{C}$, as in (\ref{eq_feedback_penalty}), (\ref{eq_feedback_maintenance}), and (\ref{eq_feedback_routing}), respectively\;
- Calculating $B$ as in (\ref{eq_convergence_1})\;

 \eIf{ $\theta_1^{*k} + \theta_2^{*k} \geq B$ }
 {
 - Convergence condition is met\;
 } 
 {
 - $k \leftarrow k+1$\;
 }
 }
 \end{minipage} }
 \vspace{0.1em}
 \caption{Decomposition algorithms, where $k$ is the iteration index.}
 \label{al_decom}

\end{algorithm}

\section{Evaluation}
\label{sec_eva}
\subsection{Parameter setting}

We implement the GADOP framework as a three-stage integer programming by using GAMS~\cite{ref_gams}. The experiments presented in this section are solved by the CPLEX solver. We use CPLEX default setting which comes with the heuristic option. 
Unless otherwise stated, we consider three drones and one truck in the GADOP framework. All the three drones are of the same type, the initial costs of which are $\widehat{c}^{(i)}_1,\widehat{c}^{(i)}_2, \widehat{c}^{(i)}_3 = S\$100$ and the capacity limit of which are $\widehat{f}_1,\widehat{f}_2,\widehat{f}_3= 2$ kilograms. The daily flying distance limit and the flying distance limit per trip are $\widehat{l}_d = 150$ kilometers and $\widehat{e}_d = 15$ kilometers for each drone, respectively. The truck is a panel van, the initial cost of which is $\bar{c}^{(i)}_1 = S\$280$ and the capacity limit of which is $\bar{f}_1 = 1,060$ kilograms. The daily distance traveling limit and daily traveling time limit are $\bar{l_t} = 200$ kilometers and $\bar{h_t}= 8$ hours, respectively. The driving speed is set at $q_{i',j'} = 50$ kilometers per hour. The initial costs of the drones and the truck are synthesized based on Singapore shipping services~\cite{ref_vehicle}. The traveling cost of the truck is approximated to $\bar{c}^{(r)}_{i',j'} = k_{i',j'} \times 1.05 \times 0.1$ based on Singapore government statistics~\cite{ref_gov}, where $1.05$ is the approximate fuel price factor and $0.1$ is the average fuel consumption factor. The cost of a drone traveling round-trip between location $i$ and the depot is $\bar{c}^{(r)}_i = 0.005 \times (k_{i,0} + k_{0,i}) $, where $0$ is the index of the depot.

Furthermore, all customers have the same package weight $g_i = 1$ kilogram. The serving time of the truck is $\bar{r} = 15$ minutes while the serving time of the drone is negligible because the drone just only drops a package off. The penalty and repair cost are $p = S\$20$ and $m = S\$50$, respectively. The penalty is set based on the cost of outsourcing package delivery to the Speedpost service~\cite{ref_singpost} by Singpost carrier plus additional cost for collecting the package or giving the compensation to a customer. Note that all customers have no specific time windows due to the space limit. Nevetheless, the experiment associated with time windows can be found in~\cite{ref_VTC1}.

We evaluate the GADOP framework using two geographically data sets, i.e., Solomon Benchmark suite C101~\cite{ref_solomon} and a real trace from one of the Singapore logistics companies. Since Singapore has many high buildings, i.e., residential buildings, we use Singapore road network as the drone flying paths. The distances in kilometers between one location to another are obtained from the Google Map~\cite{ref_googleAPI}. Solomon Benchmark suite is used in all experiments except the experiment in Section~\ref{sec_real}.

\begin{figure*}
\centering
\hspace{-1em} 
$\begin{array}{ccc}
\includegraphics[trim={5em 0 0 0},clip,width=0.35\textwidth]{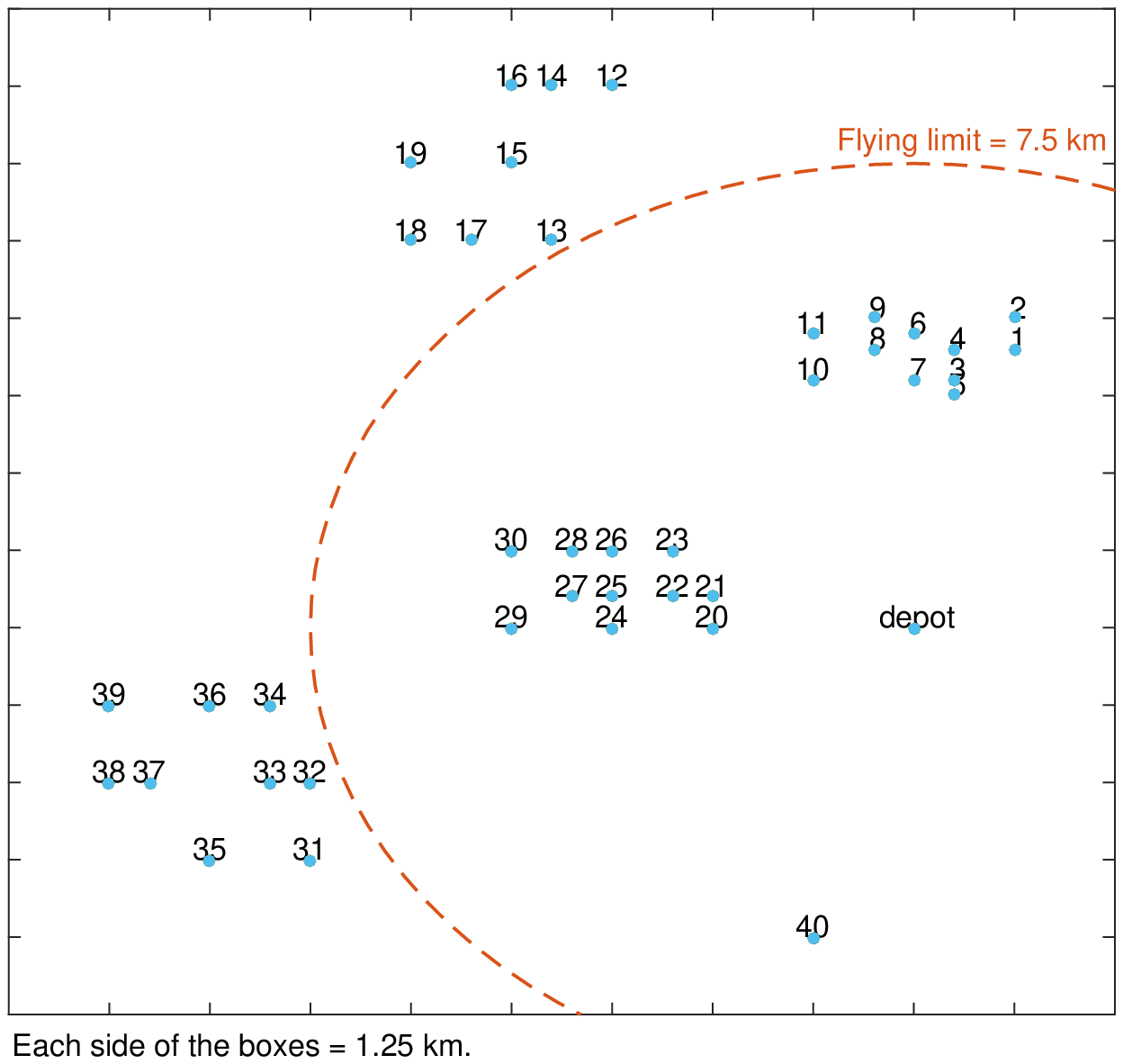} & \hspace{-2.2em} 
\includegraphics[trim={5em 0 0 0},clip,width=0.35\textwidth]{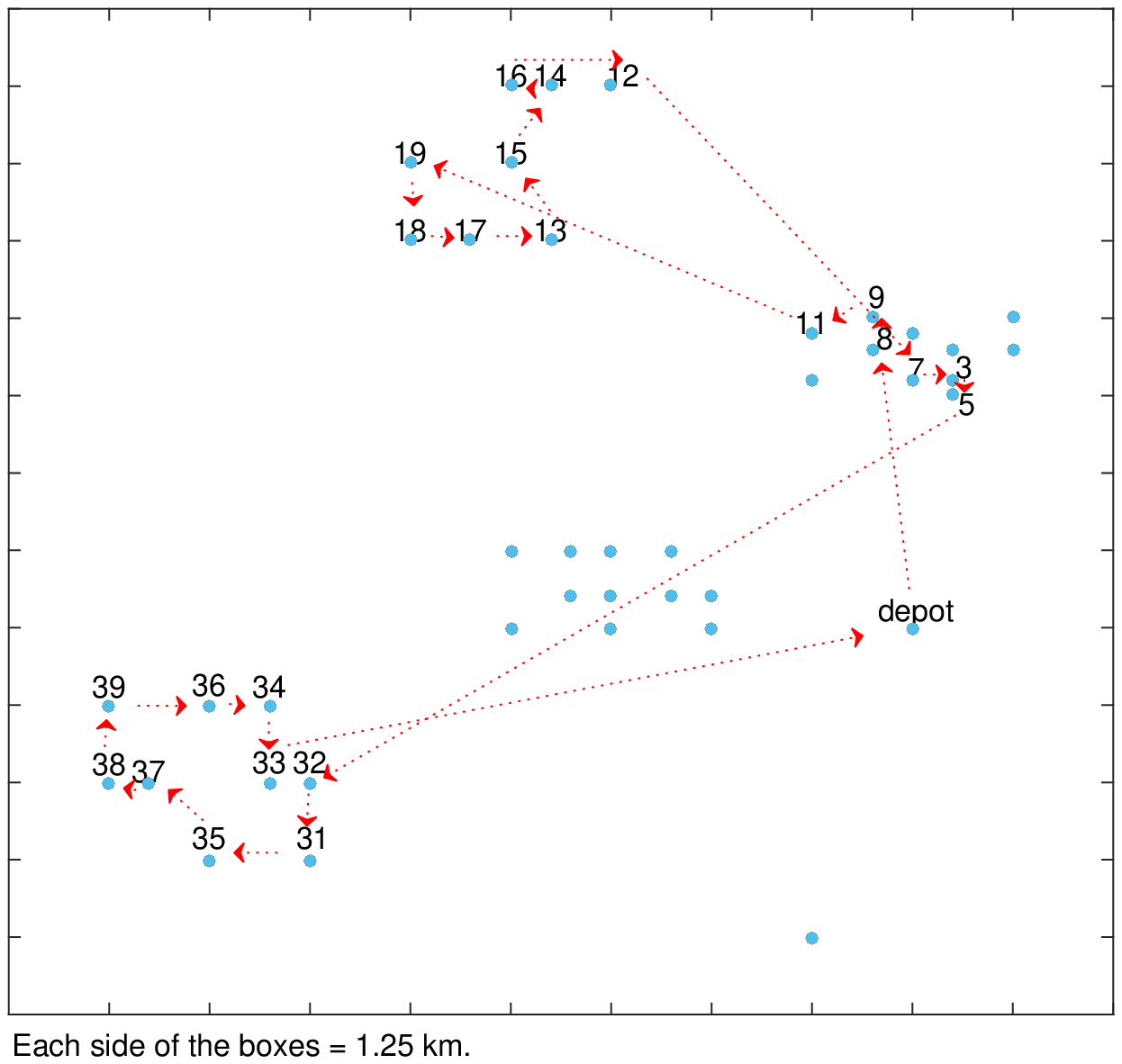} & \hspace{-2.2em}
\includegraphics[trim={5em 0 0 0},clip,width=0.35\textwidth]{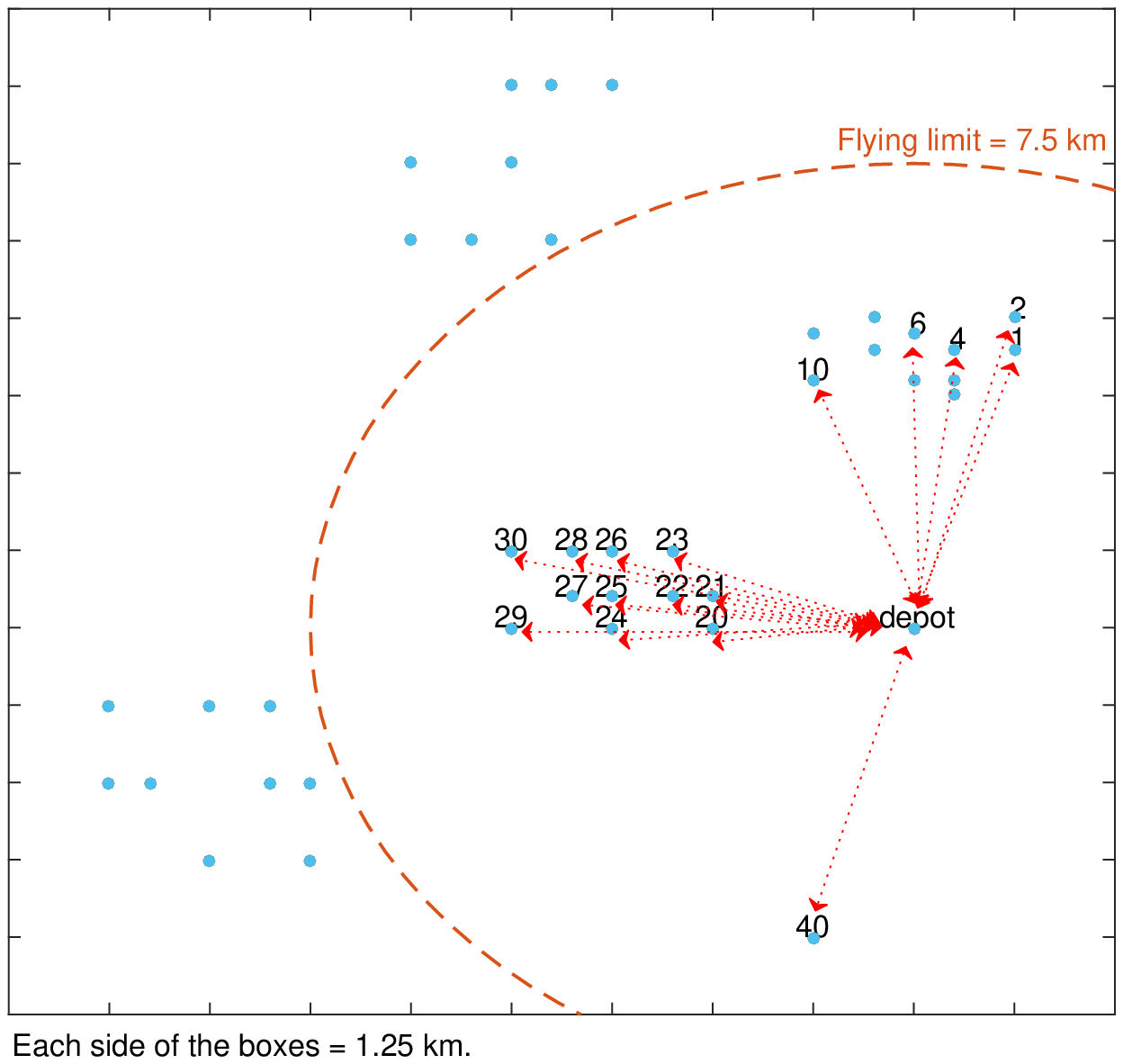} \vspace{-1.5em}\\
\vspace{-0.5em}
(a)& \hspace{-2.2em}(b) &\hspace{-2.2em} (c)
\end{array}$
 \caption{Solomon benchmark example with the scenario that the drones are always able to takeoff and no breakdown occurs, where (a) presents customers' locations, (b) presents routing path of the truck, and (c) presents the customers served by the drone. }
 \label{f_solomon_ex_nob}
\end{figure*}

\subsection{Tradeoff between taking and not-taking uncertainty into account }
\label{sec_tradeoff}

We evaluate the GADOP framework with 40 customers. The Solomon benchmark suite is tested under (i) the deterministic case where the drones are always able to takeoff and no breakdown occurs and (ii) the random case where the drones may not be able to take off and breakdown may occur.

In the deterministic case, the scenarios are fixed as $\Omega =\{\omega_1\}$, $\Lambda = \{\lambda_1\}$, and all $\mathbb{R}_d(\omega_1) =0, \mathbb{B}_{i,d}(\lambda_1) = 0$. The customer locations and the drone flying distance limit are presented in Figure~\ref{f_solomon_ex_nob}(a). 
The solution of the optimization in the deterministic case is to serve 17 and 23 customers by one drone and one truck, respectively. The routing path of the truck is presented in Figure~\ref{f_solomon_ex_nob}(b), and the customers that are served by the drone are presented in Figure~\ref{f_solomon_ex_nob}(c). The total payment is $S\$386.29$, which includes the initial cost of the truck $S\$280$, the initial cost of the drone $S\$100$, the routing cost of the truck $S\$5.59$, and the routing cost of the drone $S\$0.70$. The total traveling distances of the drone and truck are 140.27 and 53.26 kilometers, respectively.

\begin{figure*}
\centering
$\begin{array}{cc} 
\includegraphics[trim={5em 0 0 0},clip,width=0.35\textwidth]{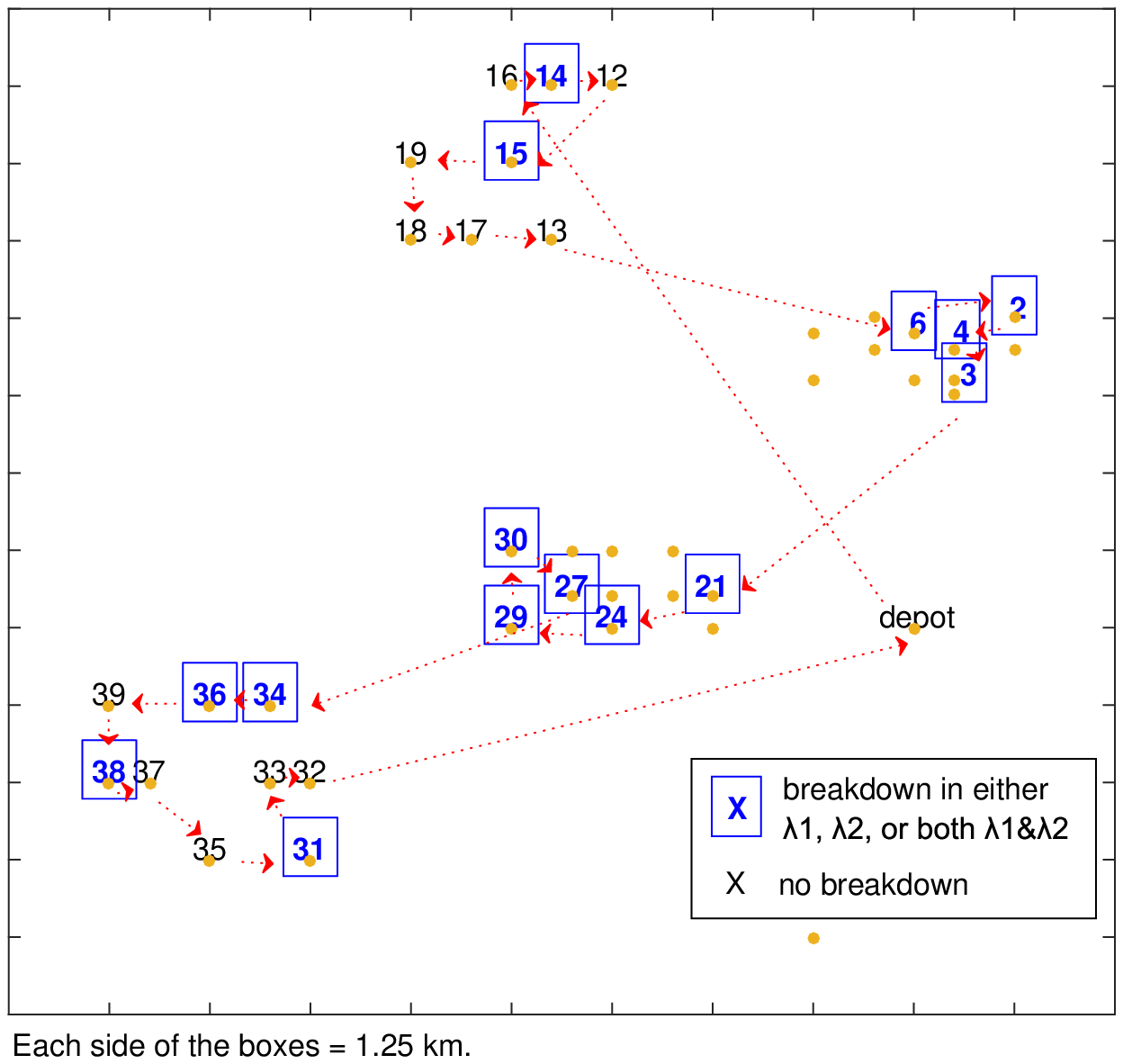} & \hspace{-2.2em} 
\includegraphics[trim={5em 0 0 0},clip,width=0.35\textwidth]{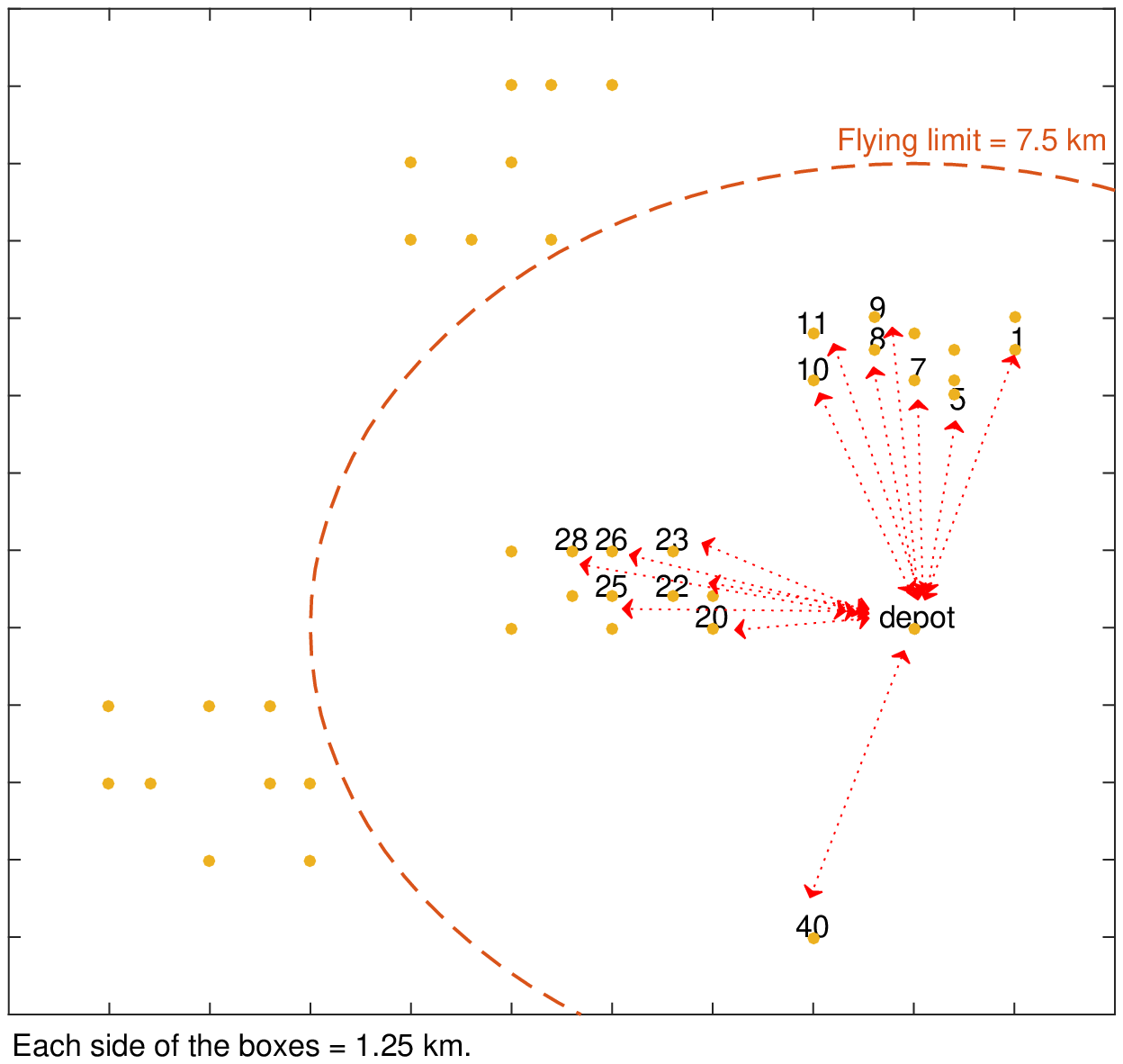} \vspace{-1.5em}
\\(a) & \hspace{-2.2em} (b) 
\end{array}$

\captionof{figure}{Solomon benchmark example with two takeoff and breakdown scenarios, where (a) presents the routing path of the truck, and (b) presents the customers served by the drone. } 
\label{f_solomon_ex_b} 

\end{figure*}

Furthermore, in the random case, there are two scenarios for each of takeoff condition and breakdown condition ($\Omega =\{\omega_1,\omega_2\}$, $\Lambda = \{\lambda_1,\lambda_2\}$). The probabilities of two takeoff scenarios are $\mathbb{P}(\omega_1)=0.9$ and $\mathbb{P}(\omega_2)=0.1$, where all drones can takeoff in scenario $\omega_1$ and cannot takeoff in scenario $\omega_2$. The probabilities of two random breakdown scenarios are also set to $\mathbb{P}(\lambda_1)=0.9$ and $\mathbb{P}(\lambda_2) =0.1$. The solution of the optimization in the random case is to serve 14 and 26 customers by one drone and one truck, respectively. The solution is presented in Figure~\ref{f_solomon_ex_b}. The total payment is $S\$414.25$ including the initial cost of the truck $S\$280$, the initial cost of the drone $S\$100$, the routing cost of the truck $S\$5.736$, the routing cost of the drone $S\$0.51$, and the penalty incurred from the takeoff scenarios $S\$28$. The total traveling distances of the drone and the truck are 114.44 and 54.63 kilometers, respectively. Obviously, only drone delivery cannot serve all customers because some customers are outside the flying area. We observe that the solution assigns the drone to serve as many customers as possible in the deterministic case because the traveling cost of the drone is cheaper. Note that daily traveling distance limit of the drone is 150 kilometer. On the contrary, in the random case, since the drone may not be able to take off or the breakdown can occur, the solution avoids the penalty and repair cost by assigning more number of customers to the truck.


\subsection{Routing, payment, and explanation of the real dataset}
\label{sec_real}
We next test the GADOP framework with the real data set from one of the Singapore logistics companies. In this experiment, we use the same takeoff and breakdown scenarios as the random case in Section~\ref{sec_tradeoff}, i.e., the two takeoff scenarios and two breakdown scenarios. 

\begin{figure*}
\center
$\begin{array}{cc}
\includegraphics[trim={0 0 15em 5em},clip,width=0.49\textwidth]{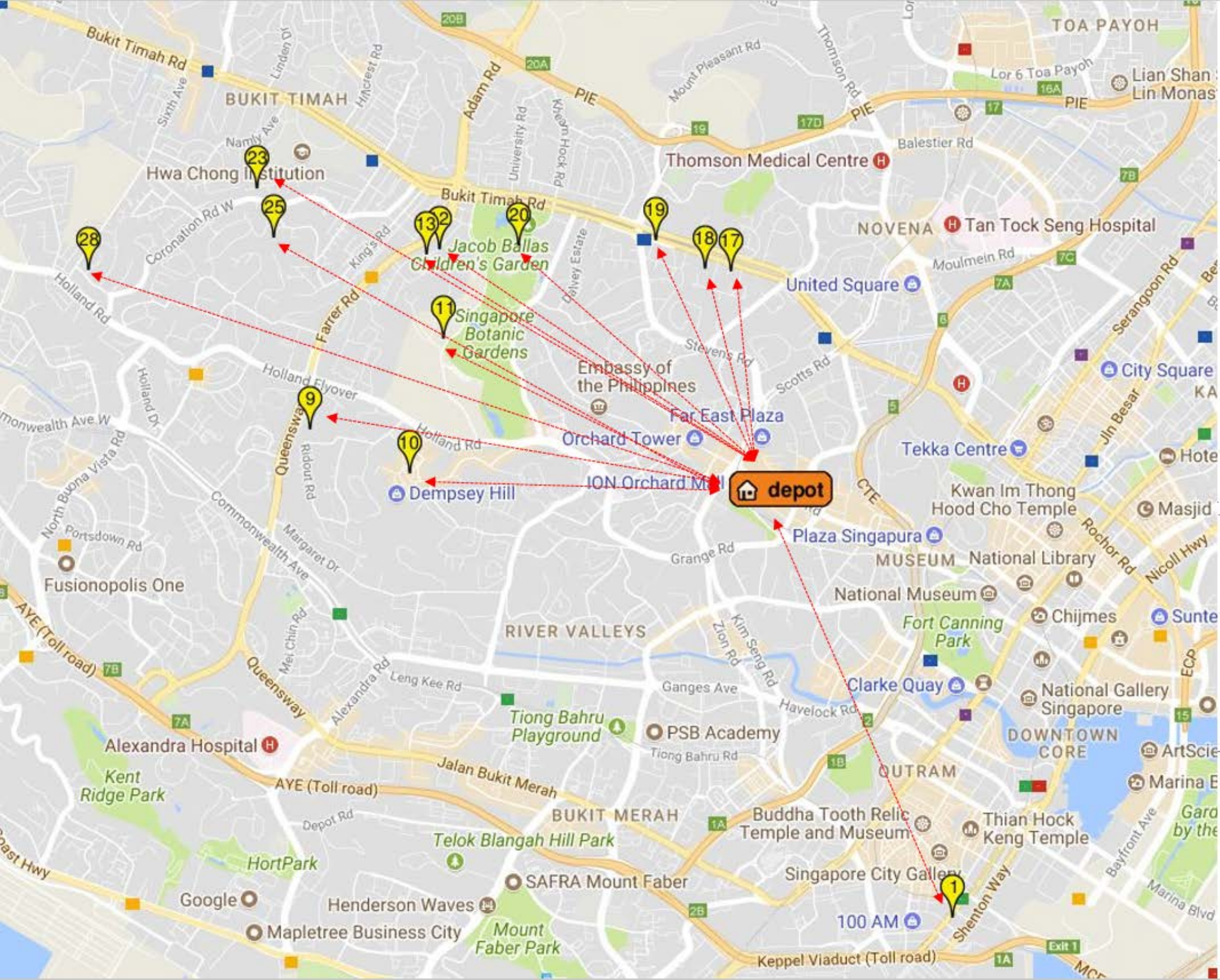} & 
\includegraphics[trim={0 0 15em 5em},clip,width=0.49\textwidth]{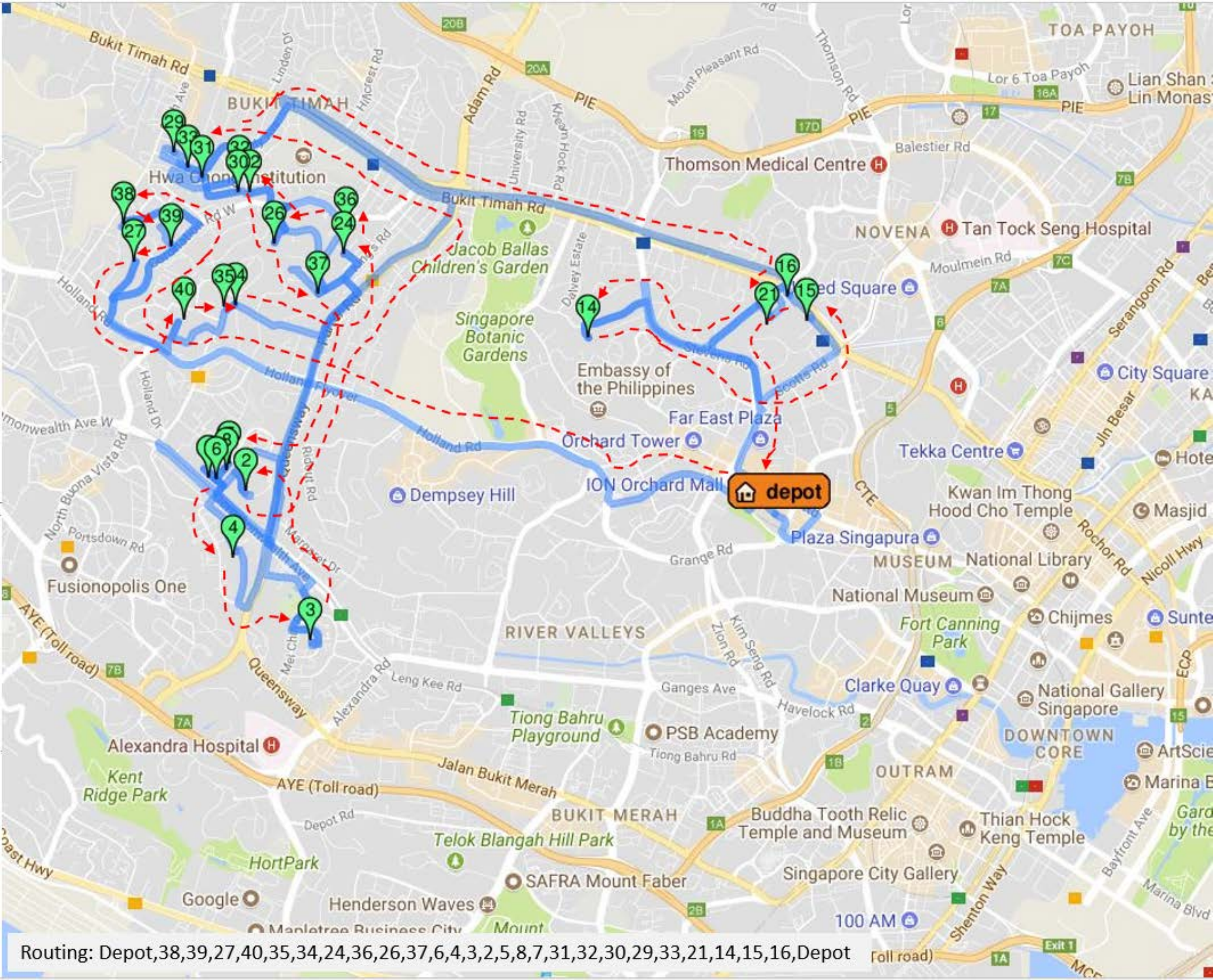} \\
(a)\text{ Drone delivery} & (b)\text{ Truck delivery}
\end{array}$
\caption{The real package delivery problem in Singapore. (a) and (b) present the customers who are served by the drone and the truck, respectively.}
 \label{f_routeMap}
\end{figure*}

The solution of the optimization is presented in Figure~\ref{f_routeMap}, which is to use one drone to serve 13 customers and one truck to serve 27 customers. The total payment is $S\$411.76$ including the initial cost of the truck $S\$280$, the initial of the drone $S\$100$, the routing cost of the truck $S\$5.17$, the routing cost of the drone $S\$0.58$, and the penalty incurred from the takeoff scenarios $S\$26$. The total traveling distances of the drone and the truck are 129.20 and 49.31 kilometers, respectively. We observe that the truck is used for the customers which are clustered in the same area or located along the route. Furthermore, the drone tends to be used for the customer far away from each other such as customer 1 at the bottom right corner of the map (Figure~\ref{f_routeMap}).

\begin{figure*}[h]
\hspace{-3em} 
\begin{minipage}{0.39\textwidth}
\includegraphics[width=\textwidth]{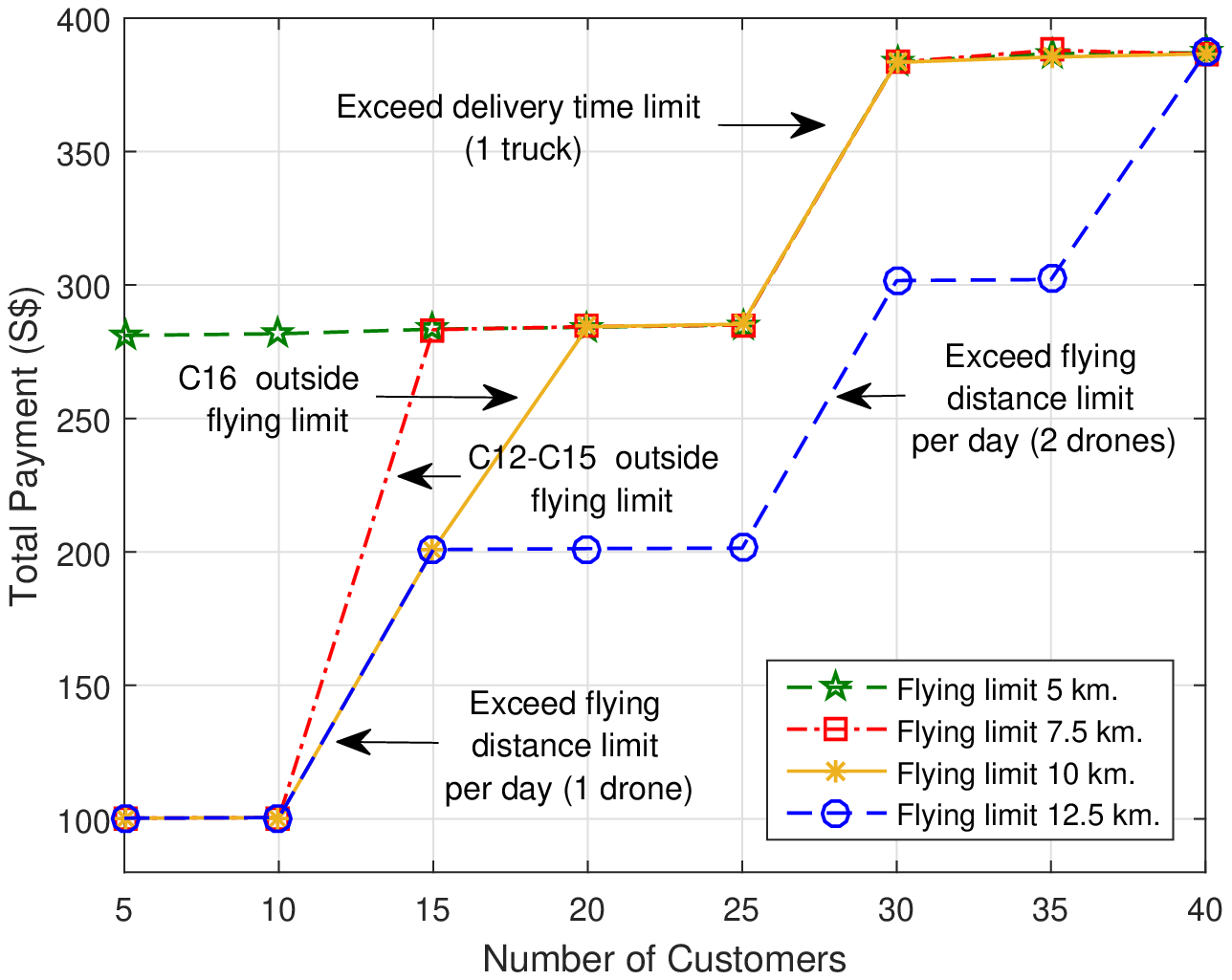}\\
\centering $(a)$
\end{minipage}
\hspace{-0.5em}
\begin{minipage}{0.35\textwidth}
\includegraphics[trim={4em 0 0 0},clip,width=\textwidth]{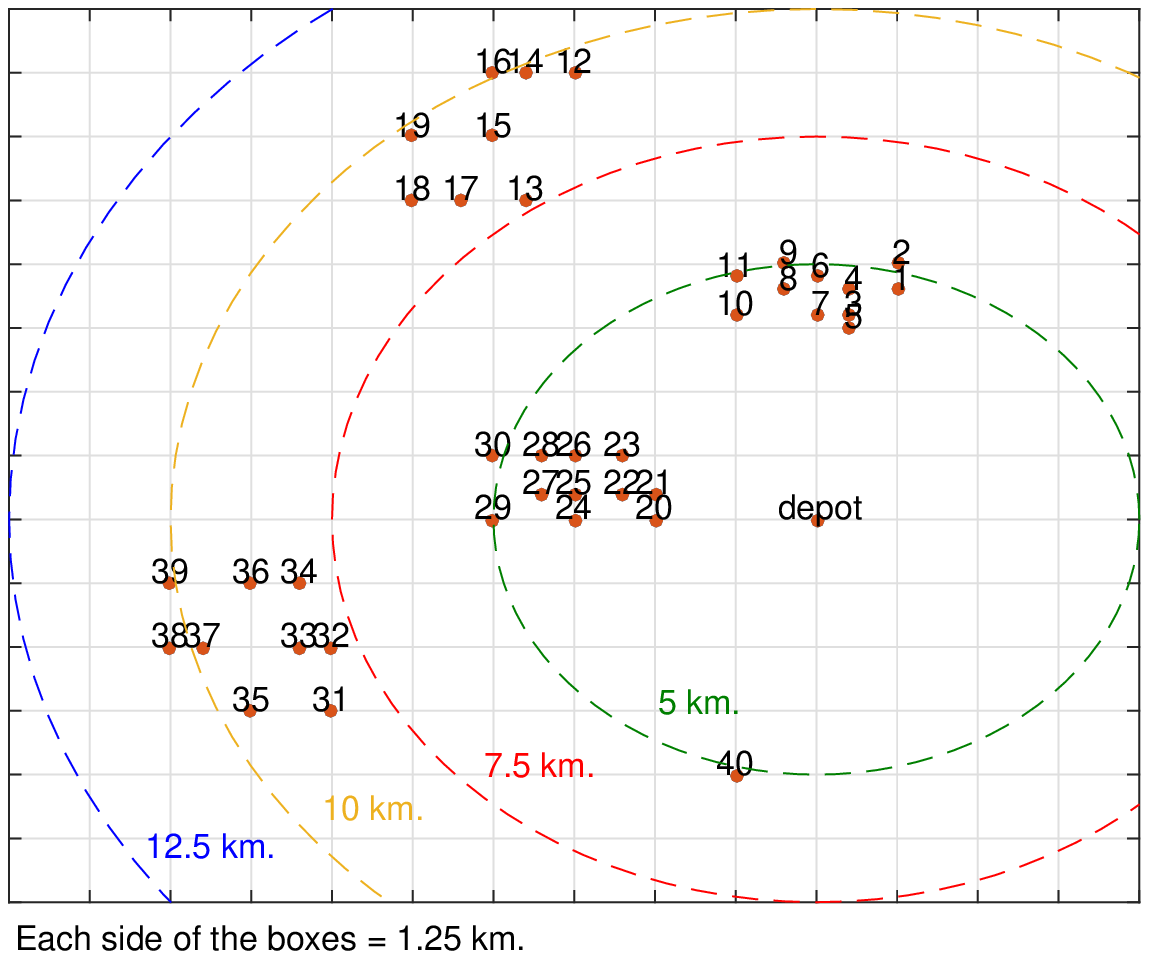}\\ \vspace{0.2em}\centering $(b)$
\end{minipage}
\hspace{-1em}\begin{minipage}{0.39\textwidth}
\includegraphics[width=\textwidth]{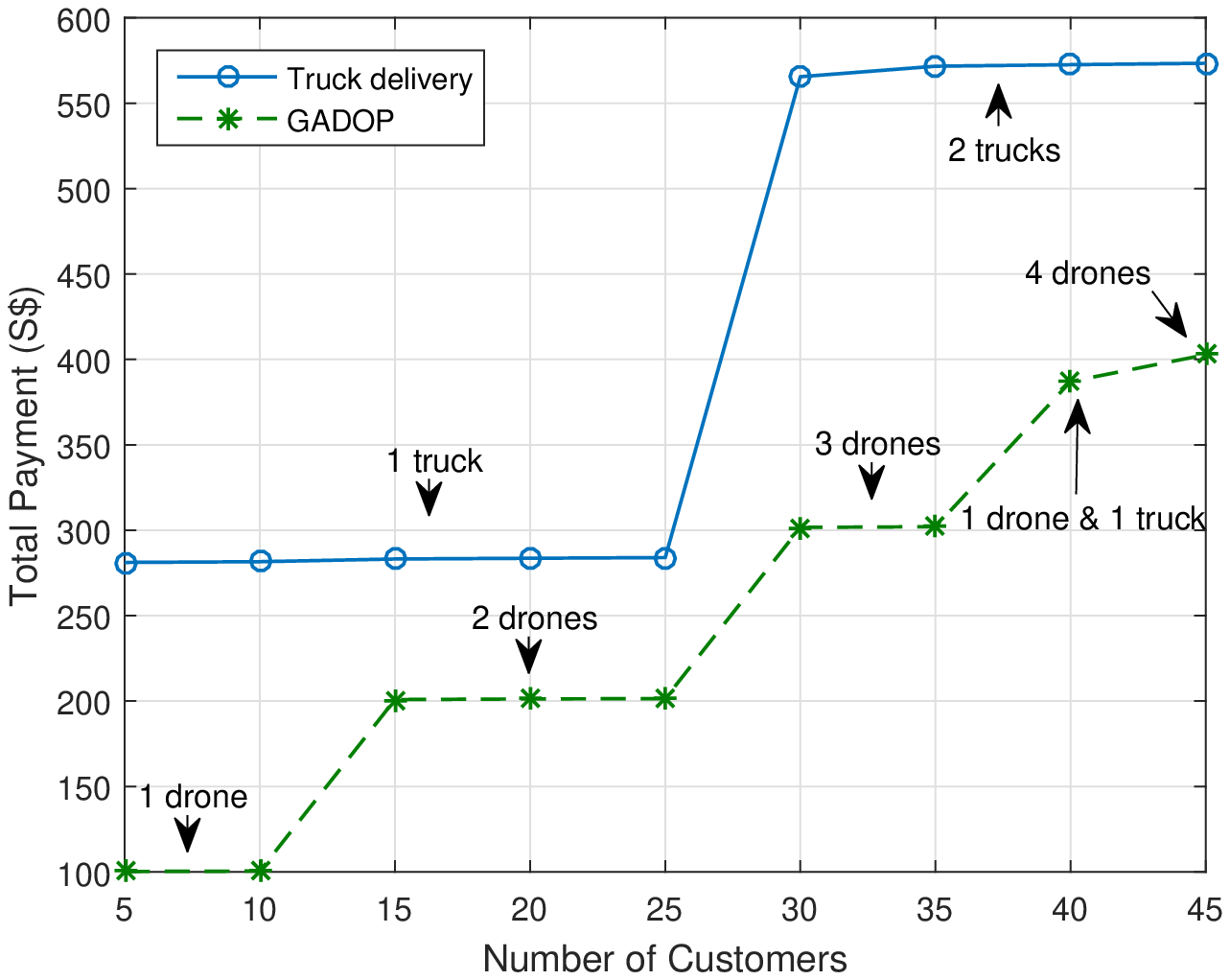}\\
\end{minipage} 
\begin{minipage}{0.66\textwidth}
\captionof{figure}{Increasing number of customers, where (a) presents the total payment and (b) presents customer locations and drone flying limits per trip. } 
\label{f_increase_customers} 
\end{minipage} \hspace{1.5em} \begin{minipage}{0.33\textwidth}
\captionof{figure}{Performance comparison between truck delivery and the proposed GADOP framework.} 
\label{f_compare}
\end{minipage}
\end{figure*}

\begin{figure*}[h]
\hspace{-3em} \begin{minipage}{0.39\textwidth}
\begin{tikzpicture}
 \draw (0, 0) node[inner sep=0] {\includegraphics[width=\textwidth]{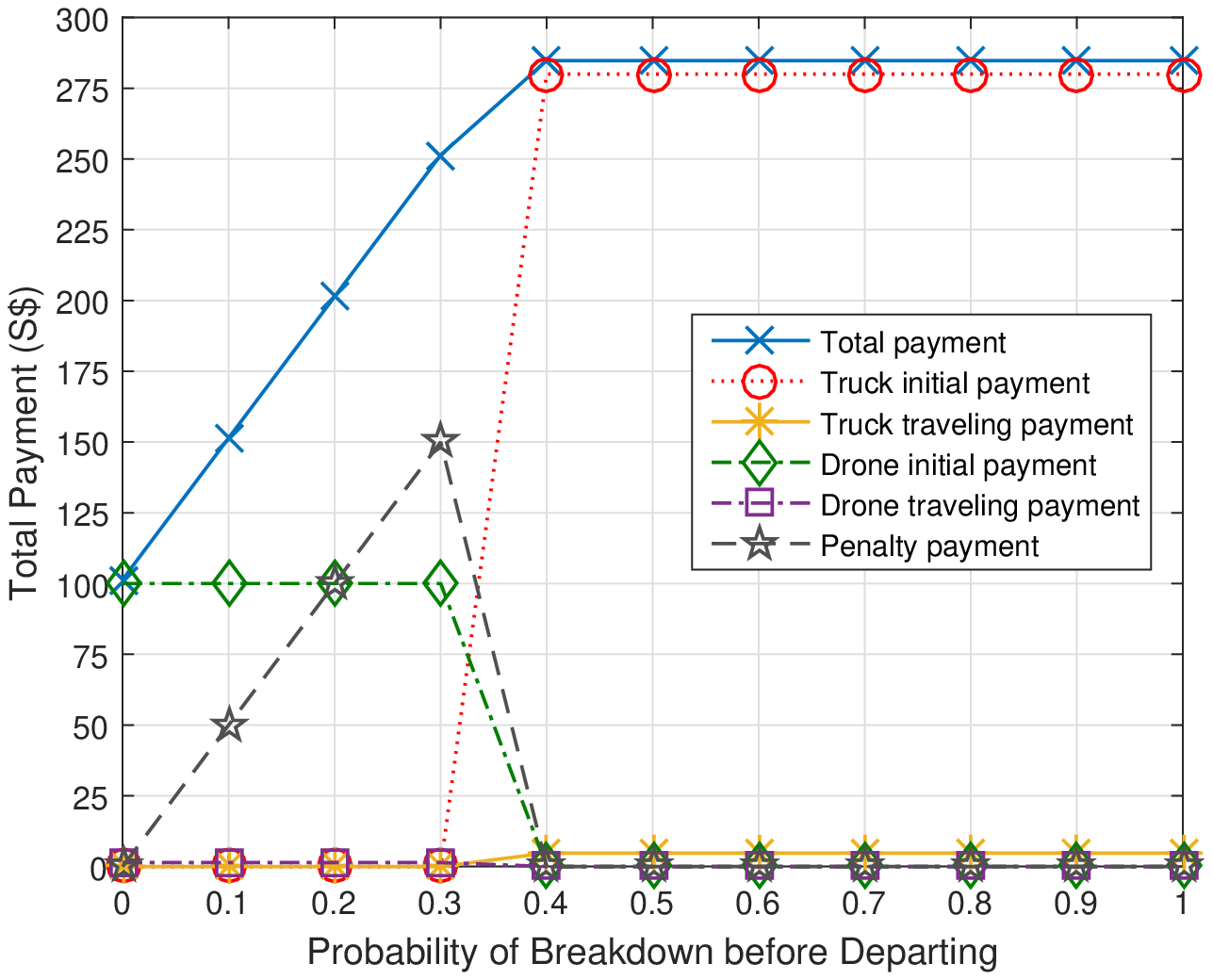} };
 \draw (0.3, -2.8) node {\scriptsize($\mathbb{P}(\omega_2)$)};
\end{tikzpicture} \\
\end{minipage}
\hspace{-2em}
\begin{minipage}{0.39\textwidth}
\begin{tikzpicture}
 \draw (0, 0) node[inner sep=0] {\includegraphics[width=\textwidth]{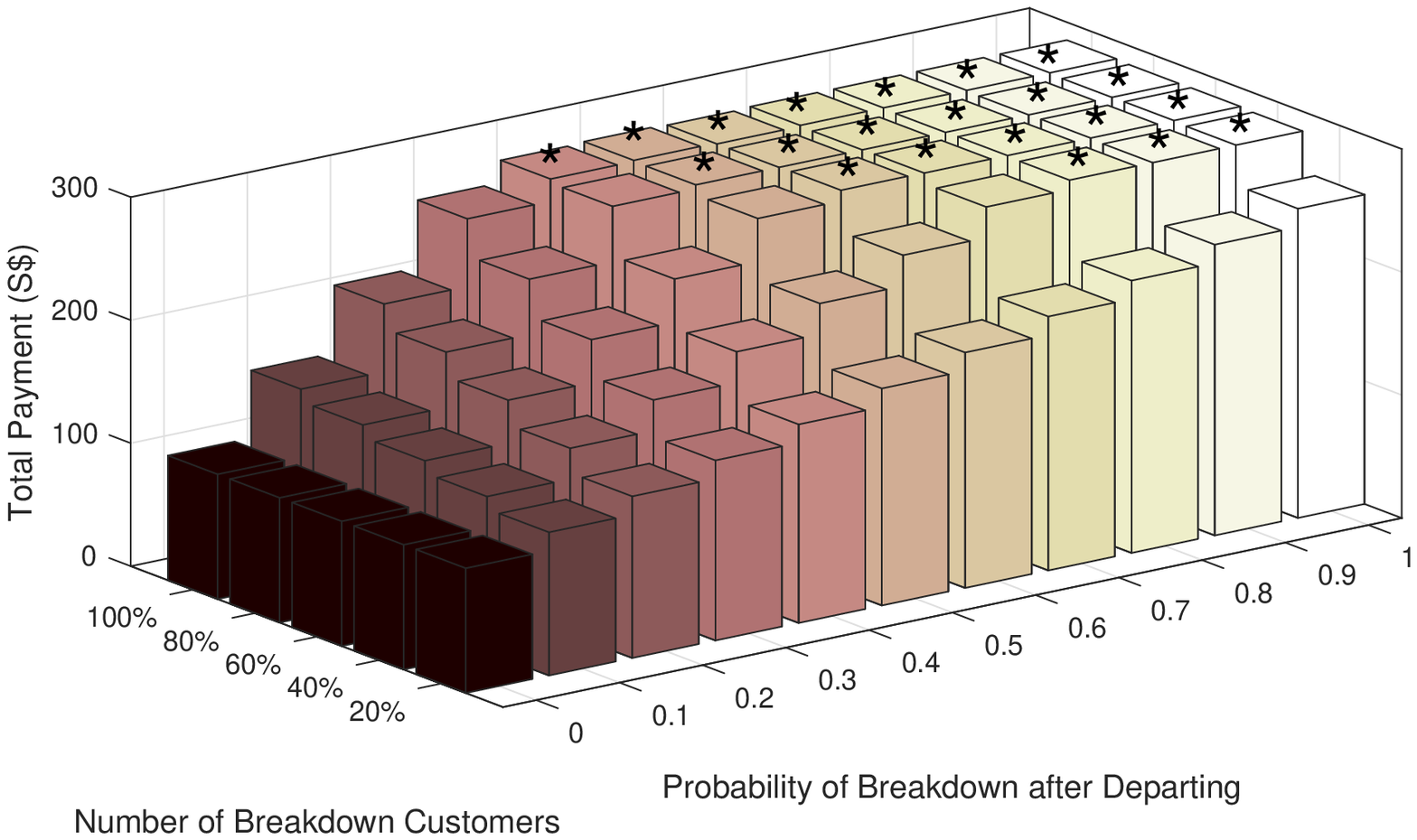} };
 \draw (-2, -2.1) node {\scriptsize($\lambda_2$)};
 \draw (1.2, -1.9) node {\scriptsize($\mathbb{P}(\lambda_2)$)};
\end{tikzpicture}
\end{minipage}
\hspace{-1em}
\begin{minipage}{0.39\textwidth}
\includegraphics[width=\textwidth]{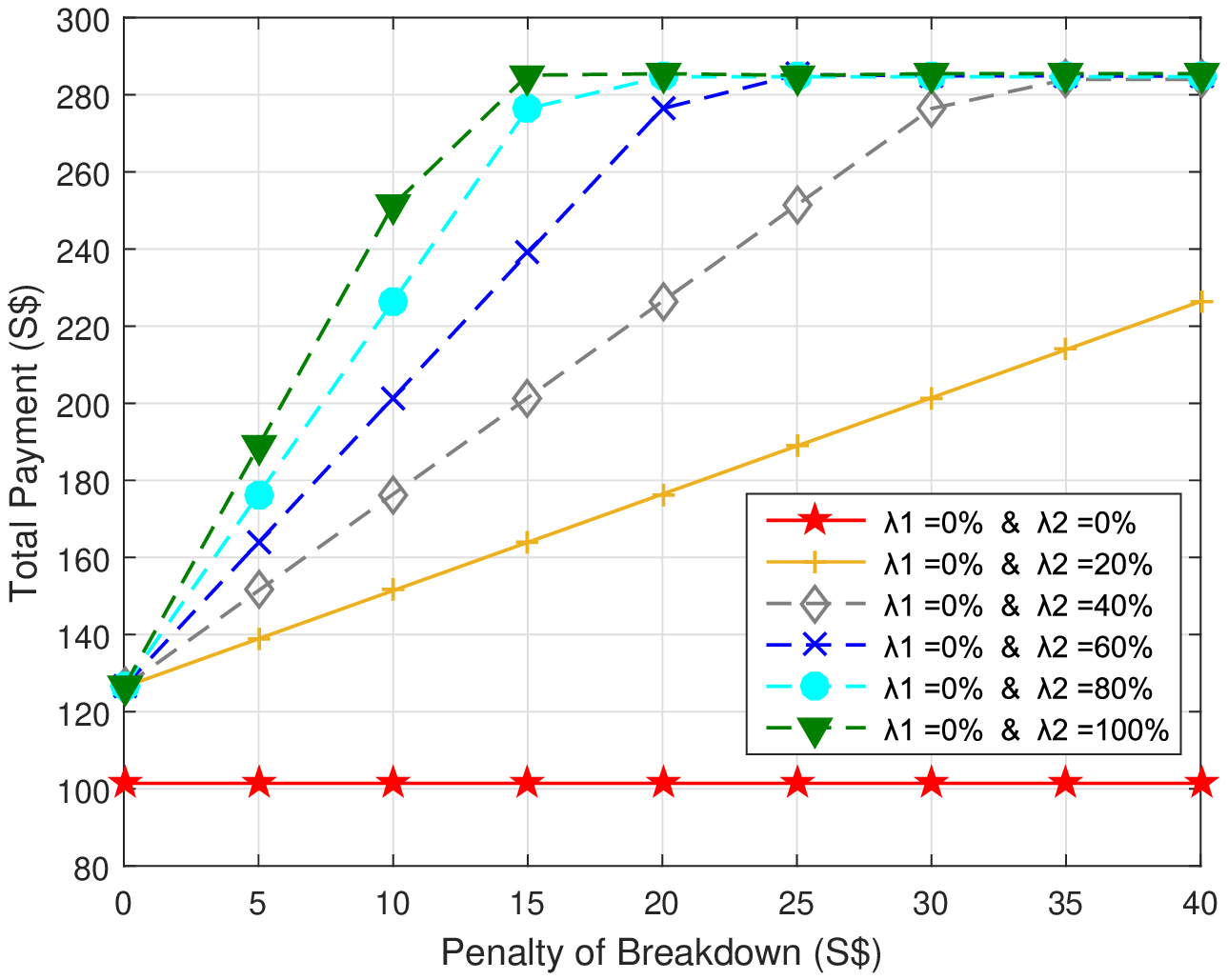}
\end{minipage} 

\hspace{-1.5em}
\begin{minipage}{0.3\textwidth}
\captionof{figure}{ Increasing the probability of that all drones cannot take off.} 
\vspace{3.5em}
\label{f_breakdown_before} 
\end{minipage} \hspace{1.5em} \begin{minipage}{0.4\textwidth}
\caption{The results of varying the number of breakdown customers and the probability of the breakdown scenarios, where the bars with an asterisk symbol on the top represents the results that all customers are served by the truck.}
 \label{f_breakdown_after_3d}
\end{minipage} \hspace{1.5em} \begin{minipage}{0.24\textwidth}
\captionof{figure}{Effects of increasing the penalty.} 
\vspace{4.5em}
\label{f_penalty} 
\end{minipage}
\end{figure*}

\begin{figure*}[h]
$\begin{array}{ccc}
\hspace{-3.1em} 
\includegraphics[width=0.39\textwidth]{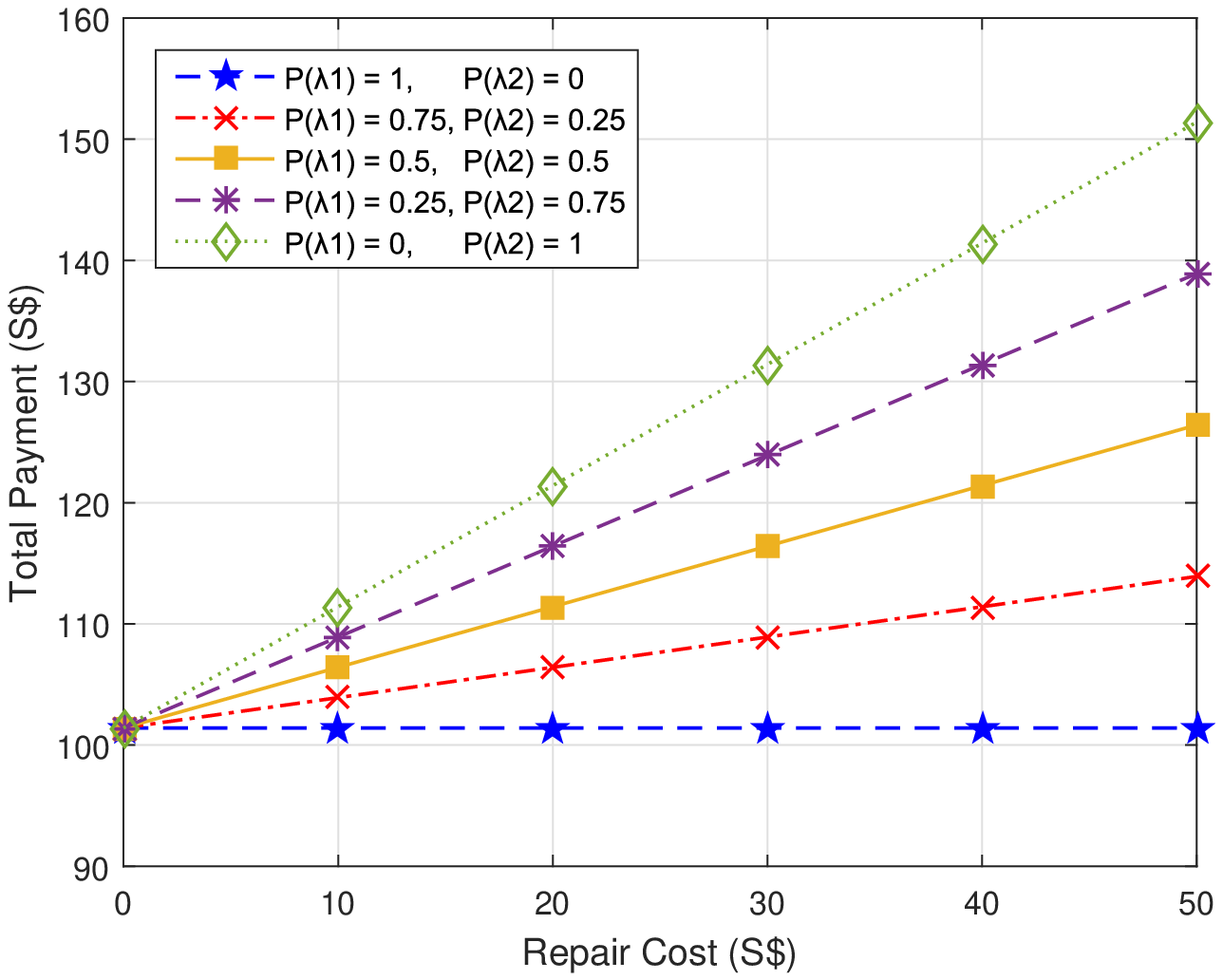} & 
\hspace{-2.8em} 
\includegraphics[width=0.39\textwidth]{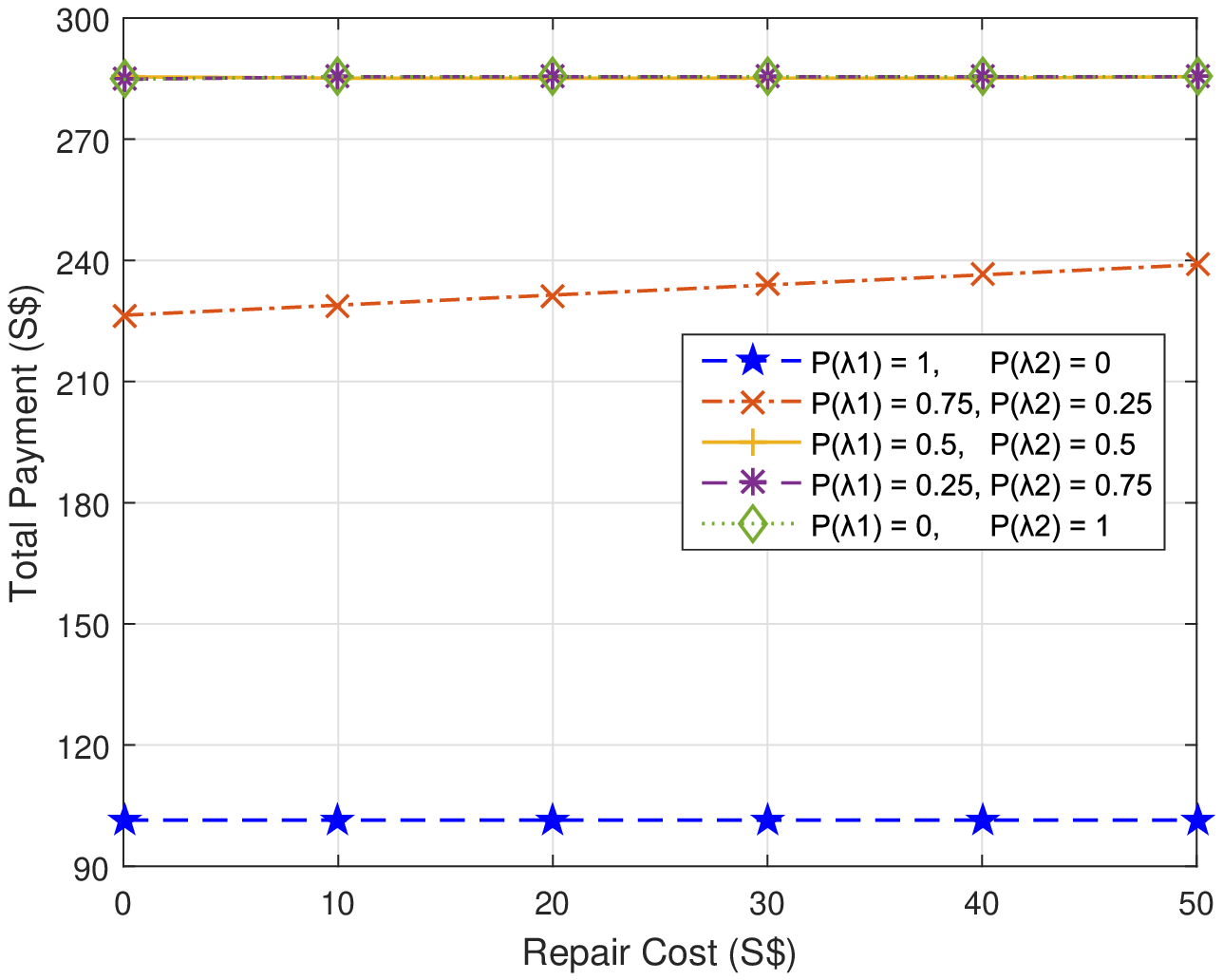} &
\hspace{-2.8em} 
\includegraphics[width=0.39\textwidth]{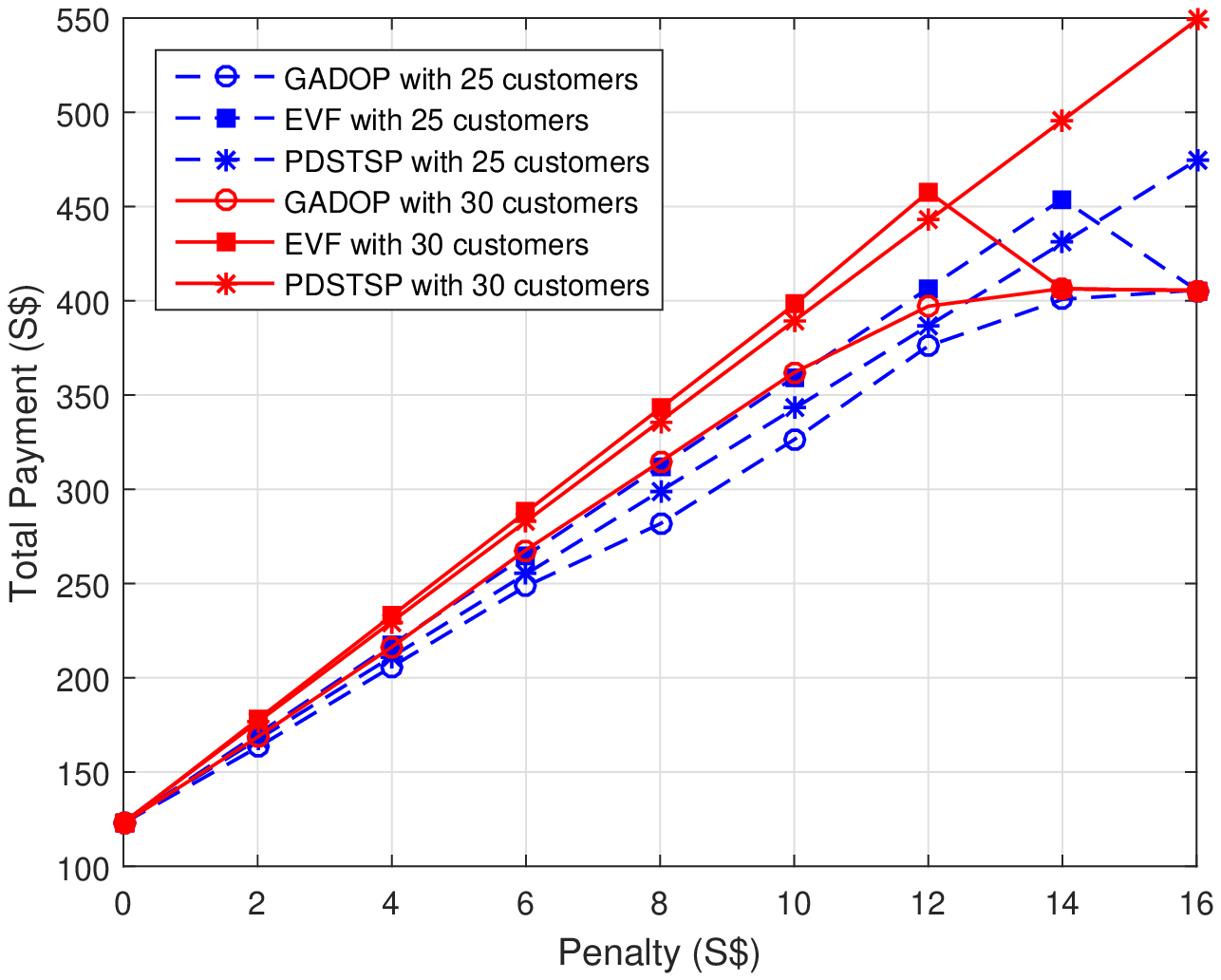} \\
\hspace{-2.8em} (a) &\hspace{-2.8em} (b) & 
\end{array}$ 

\begin{minipage}{0.66\textwidth}
\centering
\captionof{figure}{The drone repair cost is varied for (a) the penalty is $p= 0$ and (b) the penalty is $p= 20$.} 
 \label{f_repair}
\end{minipage}\hspace{1.5em}\begin{minipage}{0.33\textwidth}
\captionof{figure}{The comparison between the solutions from the GADOP, the EVF, and the PDSTSP.} 
\label{f_simulation}
\end{minipage} 
\end{figure*}

\subsection{Impacts of the number of customers and drone flying limit}

We vary the number of customers and the drone flying limit per trip. In this experiment, we test the GADOP framework with one scenario which all drones can take off and no breakdown occurs, i.e., the deterministic case. The results and the locations of the customers are presented in Figure~\ref{f_increase_customers}. If the drone flying limit per trip is large, i.e., 12.5 kilometers, all the customers are assigned to a single or multiple drones without trucks. However, when the number of the customers is large, i.e., 40, the truck and drone are used together. The reason is that three drones cannot serve all 40 customers in one day. If the drone flying limit per trip is small, i.e., 5 kilometers, the truck is selected to serve all the customers when the number of the customers is less than 25. The reason is that the drone alone cannot serve the customers farther than its flying distance limit while the truck is able to serve all the customers. When the number of the customers is larger than or equal to 30, the truck alone cannot serve all the customers because of its daily traveling time limit, and thus both the truck and drone are used together when the drone flying limit per trip is less than or equal to 10 kilometers. 
 
We conduct the performance comparison between (i) stand-alone ground-based delivery in which only ground-based trucks are used to serve customers, and (ii) joint truck and drone delivery services. As presented in Figure~\ref{f_compare}, the stand-alone ground-based delivery is always the most expensive delivery mode. This result clearly shows the effectiveness of the proposed joint truck and drone delivery.

\subsection{Impact of random takeoff scenarios}

We evaluate the GADOP framework with 25 customers, two takeoff scenarios, and one breakdown scenario, i.e., $\Lambda = \{\lambda_1\}$, and all the drones do not break down, i.e., $\mathbb{B}_{i,d}(\lambda_1) = 0$. The two takeoff scenarios are (i) all the drones can take off ($\omega_1$) and (ii) all the drones cannot take off ($\omega_2$). We vary the probability of that all the drones cannot take off, i.e., $\mathbb{P}(\omega_2)$, and $\mathbb{P}(\omega_1) = 1- \mathbb{P}(\omega_2)$. We ignore the flying distance limit in this case for the ease of presenting the impact of the takeoff scenarios. The results are presented in Figure~\ref{f_breakdown_before}. The drone is used to serve all the customers when the probability is less than or equal to 0.3. Otherwise, the truck is used to serve all the customers.

\subsection{Impact of random breakdown scenarios}
\label{sec_afterdepart}

Next we consider 25 customers, one takeoff scenario $\Omega=\{\omega_1\}$, i.e., all the drones can take off $\mathbb{R}_d(\omega_1) = 0$, and two breakdown scenarios, i.e., $\Lambda=\{\lambda_1,\lambda_2\}$. For the two breakdown scenarios, (i) there is no breakdown, i.e., all $\mathbb{B}_{i,d}(\lambda_1) = 0$, and there is breakdown for some customers. For example, breakdown percentage equal to $20\%$ means 5 customers have breakdown and 20 customers have no breakdown, which can be referred to $\mathbb{B}_{i,d}(\lambda_2) = 1$ for $i=1,\dots,5$, and $\mathbb{B}_{i,d}(\lambda_2) = 0$ for $i=6,\dots,25$. We vary the probability of the breakdown scenario $\mathbb{P}(\lambda_2)$ and the percentage of the breakdown customers. Note that we again ignore the flying distance limit. The results are presented in Figure~\ref{f_breakdown_after_3d}. From the results, the GADOP framework always selects the drone to serve the customers when the percentage of the breakdown customers is less than or equal to $20\%$ or the probability of the breakdown scenario $\mathbb{P}(\lambda_2)\leq 0.2$. 

\subsection{Impact of penalty}
\label{sec_penalty}

We use the same setting and scenarios as in Section~\ref{sec_afterdepart}, but the probabilities of the breakdown scenarios are equal, i.e., $\mathbb{P}(\lambda_1) = \mathbb{P}(\lambda_2) = 0.5$. The results are presented in Figure~\ref{f_penalty}. When the breakdown penalty cost is high, i.e., $p \geq 15$, the truck is used to serve all the customers. On the contrary, the drone is used when the penalty cost is $p < 15$. The same result is also observed when the percentage of breakdown customers is $80\%$, $60\%$, and $40\%$ when $p \geq 20, 25$, and $35$, respectively. Apparently, the number of breakdown customers has a significant impact on the solution.

\subsection{Impact of repair cost} 

Next, we consider two specific scenarios that (i) there is no breakdown and (ii) there is breakdown to all customers. We vary the drone repair cost and the probability of these two scenarios. Figure~\ref{f_repair} shows the total payments, where (a) the penalty is ignored $p=0$ and (b) the penalty is $p=20$. When the probability  that the breakdown occurs to all customers, i.e., $\mathbb{P}(\lambda_2) = 0$, the repair cost is not taken into account. As shown in Figure~\ref{f_repair}(a), the drone is used to serve all the customers. On the other hand, Figure~\ref{f_repair}(b) shows that the drone is selected when the probability that the breakdown occurs to all customers is $\mathbb{P}(\lambda_2) \leq 0.25$. Otherwise, the truck is used. We observe that the impact of the drone repair cost is significantly less than the penalty $p$ and the probability of that the breakdown occurs to all customers.

\subsection{Simulation}

We conduct a simulation and compare the performance in terms of the total payments obtained from the solutions of the GADOP optimization, expected value formulation (EVF), the parallel drone scheduling traveling salesman problem (PDSTSP)~\cite{sidekick}. We adapt the PDSTSP from minimizing the delivery time to minimizing total payments. The EVF and adapted PSDTSP formulations can be found in the Appendix. 

Here, we set the initial cost of a truck to $S\$400$ and the daily traveling time limit is 10 hours. For the EVF, all the decisions are optimized only for the first stage. For the PDSTSP, the uncertainties are not considered. Clearly, the total payment obtained from solving the GADOP optimization is significantly lower than that of the EVF and the PSDTSP. The reason is that the solution of the GADOP optimization takes into account the takeoff and breakdown conditions explicitly which lower the chance of incurring penalty, i.e., by using trucks for more customers. On the contrary, the solution of the EVF considers only the probability of unsuccessful delivery and implicitly ignores the effect of the penalty, even though the penalty is considered. Note that when the penalty is large, e.g., $S\$14$ and $S\$16$ for the cases with 30 and 25 customers, respectively, the trucks are used for all customers which is the same for both GADOP optimization and EVF. However, the PSDTSP decisions are not influenced by the penalty. Therefore, the total payments of GADOP and the EVF are identical, and the total payment of PSDTSP is the highest.

\subsection{Decomposition}

\begin{figure}
\centering
$\begin{array}{c} 
\hspace{0em} 
\includegraphics[width=0.5\textwidth]{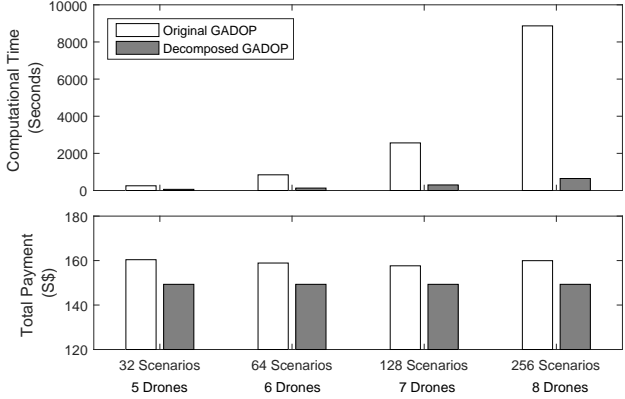} 
\end{array}$
\captionof{figure}{The comparison between the original GADOP and the decomposition. The number of takeoff scenarios and the number of drones are varied.    } 
\label{f_decom}
\end{figure}

We study the decomposition of the GADOP optimization. The penalty, repair cost, and the drone initial cost  are $p=5$, $m=5$, and $\widehat{c}^{(i)}_d =10$ respectively. In this experiment, we consider one truck and 10 breakdown scenarios ($\Lambda = \{\lambda_1,\lambda_2,\dots, \lambda_{10}\}$). The number of drones and the number of takeoff scenarios are varied. The  takeoff scenarios include all possibilities of takeoff events that can occur to all the drones. To illustrate,  five drones are experimented with $32$ takeoff scenarios ($\Omega = \{\omega_1,\omega_2,\dots, \omega_{32}\}$), six drones are experimented with $64$ takeoff scenarios, and so on. 

The comparison is shown in Figure~\ref{f_decom}. When the number of takeoff scenarios grows, the difference between the computational time of the original and decomposed problems becomes larger. As shown in Figure~\ref{f_decom}, the decomposed GADOP achieves $407\%$, $648\%$, $854\%$, and $1372\%$ speedup with 32, 64, 128, and 256 scenarios, respectively. Meanwhile, the decomposed GADOP always achieves lower total payment. Note that, in some cases,  the CPLEX solver cannot give  an optimal solution for the original problem when the problem is too large. Thus, the CPLEX solver may return a suboptimal solution instead of the optimal solution. In this regard, the suboptimal solution achieves much higher payment than that of the decomposition while the computational time of the decomposition may not significantly improve~\cite{ref_VTC2}.   In summary, the decomposition enables the GADOP to handle larger number of scenarios.

\section{Conclusion and Future works}
\label{sec_con}

We have proposed the joint ground and aerial delivery service optimization and planning (GADOP) framework. The uncertainty of drone delivery has been taken into account by considering takeoff and breakdown conditions. The aim of the GADOP framework is to help a supplier to make the decision either to use ground-based trucks, drones, or the combination to serve customers. We have formulated the GADOP optimization as a three-stage stochastic integer programming model. The performance evaluation of both Solomon benchmark suite and the Singapore real road network has been conducted. The results clearly show that using drones in the package delivery can reduce the total payment of the supplier significantly. Moreover, simulation results have shown that the GADOP framework always achieves the lower total payment than that of the expected value formulation (EVF) and the Parallel Drone Scheduling Traveling Salesman Problem (PDSTSP). We have adopted the L-shape decomposition method to enhance the solvability of the optimization problem. 

For future work, the uncertainty in customers' demand and traveling time will be considered and, we will also incorporate multiple-stage scenarios of the breakdown condition, which depend on each customer. Moreover,  different types of subtour elimination, which aim to speedup computational time, will be studied. 

\section{Appendix}
\subsection{Expected Value Formulation (EVF)}
The EVF objective function is similar to the original objective function in~(\ref{eq_obj1}), but only the formulation of the $\mathbb{E}(\mathscr{L}(\widehat{X}_{i,d}))$ is different. The equation is indicated in~(\ref{eq_eva}), where $\mathbb{P}(d)$, $\mathbb{P}(i)$, and $\mathbb{M}(d)$ denote the probability that drone $d$ cannot take off, the probability that breakdown occurs while serving customer $i$, and the probability that drone $d$ needs to be repaired, respectively. The EVF subjects to the constraints in (\ref{con_ini_ftl})-(\ref{con_subtour}) and (\ref{eq_con_order3})-(\ref{eq_con_lastbound}). 

\begin{align}
\mathbb{E}(\mathscr{L}(\widehat{X}_{i,d})) = &
\sum_{i \in \mathcal{C}}\sum_{d \in \mathcal{D}}\widehat{c}^{(r)}_{i}\widehat{X}_{i,d}\mathbb{P}(d)\nonumber +\sum_{d \in \mathcal{D}} p\widehat{X}_{i,d}(1-\mathbb{P}(d)) \nonumber\\
&+ \sum_{i \in \mathcal{C} }\sum_{d \in \mathcal{D} }p\widehat{X}_{i,d}(1-\mathbb{P}(d))\mathbb{P}(i)\nonumber\\
&+ \sum_{d \in \mathcal{D}} m(1-\mathbb{P}(d))\mathbb{M}(i) 
\label{eq_eva}
\end{align}

\subsection{Parallel Drone Scheduling Traveling Salesman Problem (PDSTSP)}

The PDSTSP can handle multiple drones with only one truck. It does not take the serving time constraint and the constraints of traveling limits ,i.e., time and distance, into account. Furthermore, the the uncertainties i.e., takeoff and breakdown conditions, are ignored in the PDSTSP. 
We adapt the PDSTSP objective function, which is proposed in~\cite{sidekick}, from minimizing the delivery time to minimizing the total payment. The new objective function is presented in (\ref{eq_psdtsp}). The two delivery time constraints in~\cite{sidekick} are replaced by the constraints in (\ref{con_ini_psd_ftl}) and (\ref{con_ini_psd_uav}). The constraint in (\ref{con_ini_psd_ftl}) ensures that the initial cost of the truck is paid when the truck is used. Similarly, the constraint in (\ref{con_ini_psd_uav}) ensures that the initial cost of drones are paid when the drones are used.  The rest of the constraints in~\cite{sidekick} are kept as the original.  

\begin{align}
& \mbox{Minimize:} \nonumber \\
& \bar{c}^{(i)} \bar{W} + 
\sum_{d \in \mathcal{D}}\widehat{c}^{(i)}_d \widehat{W}_d + \sum_{\substack{i', j'\in \\ \mathcal{C}\cup\{0\}}}\bar{c}^{(r)}_{i',j'}V_{i',j'} + 
\sum_{i \in \mathcal{C}}\sum_{d \in \mathcal{D}}\widehat{c}^{(r)}_{i}\widehat{X}_{i,d} 
\label{eq_psdtsp}
\end{align}
\begin{align}
&\sum_{i \in \mathcal{C}} V_{i} \leq \Delta \bar{W}, & \forall i \in \mathcal{C} \label{con_ini_psd_ftl}\\
&\sum_{i \in \mathcal{C}} \widehat{X}_{i,d} \leq \Delta \widehat{W}_d, & \forall d \in \mathcal{D}\label{con_ini_psd_uav}
\end{align}

\section{Acknowledgment}
 This work was partially supported by Singapore Institute of Manufacturing Technology-Nanyang Technological University (SIMTech-NTU) Joint Laboratory and Collaborative research Programme on Complex Systems.

\newpage
\begin{IEEEbiography}
    [{\includegraphics[width=1in,height=1.25in,clip,keepaspectratio]{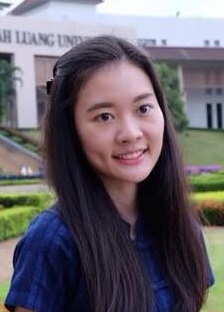}}]{Suttinee Sawadsitang} received her B.Eng in Computer Engineering from  King Mongkuts University of Technology Thonburi (KMUTT), Thailand in 2012 and  M.Eng from Shanghai Jiao Tong University, China in 2015. She is currently pursuing a Ph.D. degree at  SIMTech-NTU Joint Lab on Complex
Systems, Nanyang Technological University, Singapore. Her research interests are in the area of transportation systems, operations research, optimization, and high performance computing.

\end{IEEEbiography}

\begin{IEEEbiography}
    [{\includegraphics[width=1in,height=1.25in,clip,keepaspectratio]{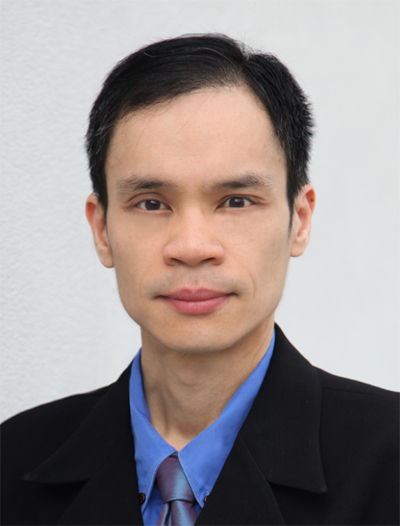}}]{Dusit Niyato}
(M'09-SM'15-F'17) is currently a professor in the School of Computer Science and Engineering, at Nanyang Technological University, Singapore. He received B.Eng. from King Mongkuts Institute of Technology Ladkrabang (KMITL), Thailand in 1999 and Ph.D. in Electrical and Computer Engineering from the University of Manitoba, Canada in 2008. His research interests are in the area of energy harvesting for wireless communication, Internet of Things (IoT) and sensor networks.
\end{IEEEbiography}

\begin{IEEEbiography}
 [{\includegraphics[width=1in,height=1.25in,clip,keepaspectratio]{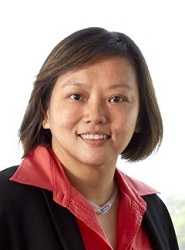}}]
{Puay Siew Tan} received the Ph.D. degree in computer science from the School of Computer Engineering, Nanyang Technological University, Singapore. She is presently an Adjunct Associate Professor with the School of Computer Science and Engineering, Nanyang Technological University and also the Co-Director of the SIMTECH-NTU Joint Laboratory on Complex Systems. In her full-time job at Singapore Institute of Manufacturing Technology (SIMTech), she leads the Manufacturing Control Tower\textsuperscript{TM} (MCT\textsuperscript{TM}) as the Programme Manager. She is also the Deputy Division Director of the Manufacturing System Division. Her research interests are in the cross-field disciplines of Computer Science and Operations Research for virtual enterprise collaboration, in particular sustainable complex manufacturing and supply chain operations in the era of Industry 4.0.
\end{IEEEbiography}

\begin{IEEEbiography}
 [{\includegraphics[width=1in,height=1.25in,clip,keepaspectratio]{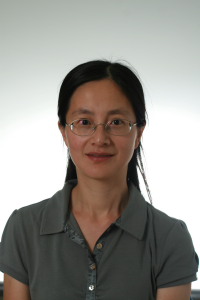}}]
{Ping Wang} (M’08, SM’15) received the PhD degree in electrical engineering from University of Waterloo, Canada, in 2008. She was with Nanyang Technological University, Singapore. Currently she is an associate professor at the department of Electrical Engineering and Computer Science, York University, Canada. Her current research interests include resource allocation in multimedia wireless networks, cloud computing, and smart grid. She was a corecipient of the Best Paper Award from IEEE Wireless Communications and Networking Conference (WCNC) 2012 and IEEE International Conference on Communications (ICC) 2007. She served as an Editor of IEEE Transactions on Wireless Communications, EURASIP Journal on Wireless Communications and Networking, and International Journal of Ultra Wideband Communications and Systems.
\end{IEEEbiography}

\end{document}